\documentclass[10pt,twocolumn,letterpaper]{article}

\usepackage{cvpr}              

\usepackage{pifont}
\usepackage{graphicx} 
\usepackage{amsmath}
\usepackage{amssymb}
\usepackage{adjustbox}
\usepackage{multirow}
\usepackage{stfloats}

\usepackage{pgfplots}
\pgfplotsset{compat=1.18}

\usepackage[linesnumbered,ruled,vlined]{algorithm2e}
\DontPrintSemicolon
\renewcommand{\KwSty}[1]{\textnormal{\textcolor{blue!90!black}{\ttfamily\bfseries #1}}\unskip}

\SetKwComment{Comment}{\color{green!50!black}// }{}

\newcommand{\assign}{\leftarrow}
\newcommand{\var}{\texttt}
\newcommand{\FuncCall}[2]{\texttt{\bfseries #1(#2)}}
\SetKwProg{Function}{function}{}{}

\SetKwInput{Input}{input}
\SetKwInput{Output}{output}

\definecolor{MyGreen}{RGB}{0, 180, 0}
\definecolor{MyRed}{RGB}{180, 0, 0}
\definecolor{MyYellow}{RGB}{180, 180, 0}
\newcommand{\cmark}{{\textcolor{MyGreen}{\ding{51}}}}%
\newcommand{\mmark}{{\textcolor{MyYellow}{(\ding{51})}}}%
\newcommand{\xmark}{{\textcolor{MyRed}{\ding{55}}}}%
\newcommand{\qmark}{{\textcolor{MyYellow}{\textbf{?}}}}%

\def\ours{ALPINE\xspace}

\newif\ifshowedits

\newcommand{\addeditor}[3]{%
  \definecolor{#1color}{rgb}{#3}
  \expandafter\newcommand\csname #1\endcsname[1]{%
  \ifshowedits
    {\color{#1color} ##1}%
  \else
    {##1}%
  \fi
  }%
  \expandafter\newcommand\csname #1rmk\endcsname[1]{%
  \ifshowedits
    {\color{#1color} {\bf [#2: ##1]}}
  \fi
  }%
  \expandafter\newcommand\csname #1rpl\endcsname[2]{%
  \ifshowedits
    {\color{#1color} ##1 \sout{##2}}
  \else
    {##1}
  \fi
  }%
}

\usepackage{nicematrix}
\usepackage{colortbl}
\newcolumntype{a}{>{\columncolor{blue!15}}c}

\newcolumntype{R}[2]{%
    >{\adjustbox{angle=#1,lap=1.3\width-(#2)}\bgroup}%
    l%
    <{\egroup}%
}

\newcommand\blfootnote[1]{%
  \begingroup
  \renewcommand\thefootnote{}\footnote{#1}%
  \addtocounter{footnote}{-1}%
  \endgroup
}

\makeatletter
\renewcommand\paragraph{\@startsection
    {paragraph}{4}{\z@}%
    {0.3ex \@plus0.5ex \@minus.2ex}%
    {-1em}%
    {\normalfont\normalsize\bfseries}}
\makeatother

\addeditor{vincent}{VL}{0.0, 0.5, 0.0}
\addeditor{gilles}{GP}{0.8, 0.5, 0.5}
\addeditor{renaud}{RM}{0.95, 0.55, 0.15}
\addeditor{alex}{AB}{0.6, 0.4, 1.0}
\addeditor{corentin}{CS}{1.0, 0.0, 1.0}
\showeditsfalse

\definecolor{cvprblue}{rgb}{0.21,0.49,0.74}
\usepackage[pagebackref,breaklinks,colorlinks,allcolors=cvprblue]{hyperref}
\usepackage{caption}

\def\paperID{151} 
\def\confName{3DV\xspace}
\def\confYear{2026\xspace}

\title{{Is clustering enough for LiDAR instance segmentation?\\A state-of-the-art training-free baseline}}

\author{
Corentin Sautier$^{1,2}$ \quad
Gilles Puy$^{2}$ \quad
Alexandre Boulch$^{2}$ \quad
Renaud Marlet$^{1,2}$ \quad
Vincent Lepetit$^{1}$ \quad
\and
\and
\large
\hspace{-3mm} \textsuperscript{1}LIGM, Ecole des Ponts, Univ Gustave Eiffel, CNRS,  France
\hspace{1mm} \textsuperscript{2}Valeo.ai, France 
}

\begin{document}
\Crefname{algorithm}{Algorithm}{Algorithms}
\crefname{algorithm}{Alg.}{Algs.}
\maketitle

\begin{abstract}
Panoptic segmentation of LiDAR point clouds is fundamental to outdoor scene understanding, with autonomous driving being a primary application. While state-of-the-art approaches typically rely on end-to-end deep learning architectures and extensive manual annotations of instances, the significant cost and time investment required for labeling large-scale point cloud datasets remains a major bottleneck in this field.
In this work, we demonstrate that competitive panoptic segmentation can be achieved using only semantic labels, with instances predicted without any training or annotations. 
Our method outperforms {most} state-of-the-art supervised methods on standard benchmarks including SemanticKITTI and nuScenes, and outperforms every publicly available method on SemanticKITTI as a drop-in instance head replacement, while running in real-time on a single-threaded CPU and requiring no instance labels.
It is fully explainable, and requires no learning or parameter tuning.
\ours combined with state-of-the-art semantic segmentation ranks first on the official panoptic segmentation leaderboard of SemanticKITTI.
\blfootnote{Code is available at \href{https://github.com/valeoai/Alpine/}{https://github.com/valeoai/Alpine/}}
\end{abstract}

\section{Introduction}
\label{sec:intro}

\begin{figure}
    \centering
    \def\svgwidth{\linewidth}
\begingroup%
  \makeatletter%
  \providecommand\color[2][]{%
    \errmessage{(Inkscape) Color is used for the text in Inkscape, but the package 'color.sty' is not loaded}%
    \renewcommand\color[2][]{}%
  }%
  \providecommand\transparent[1]{%
    \errmessage{(Inkscape) Transparency is used (non-zero) for the text in Inkscape, but the package 'transparent.sty' is not loaded}%
    \renewcommand\transparent[1]{}%
  }%
  \providecommand\rotatebox[2]{#2}%
  \newcommand*\fsize{8pt\relax}%
  \newcommand*\lineheight[1]{\fontsize{\fsize}{#1\fsize}\selectfont}%
  \ifx\svgwidth\undefined%
    \setlength{\unitlength}{866.0668657bp}%
    \ifx\svgscale\undefined%
      \relax%
    \else%
      \setlength{\unitlength}{\unitlength * \real{\svgscale}}%
    \fi%
  \else%
    \setlength{\unitlength}{\svgwidth}%
  \fi%
  \global\let\svgwidth\undefined%
  \global\let\svgscale\undefined%
  \makeatother%
  \begin{picture}(1,1.03858043)%
    \lineheight{1}%
    \setlength\tabcolsep{0pt}%
    \put(0,0){\includegraphics[width=\unitlength,page=1]{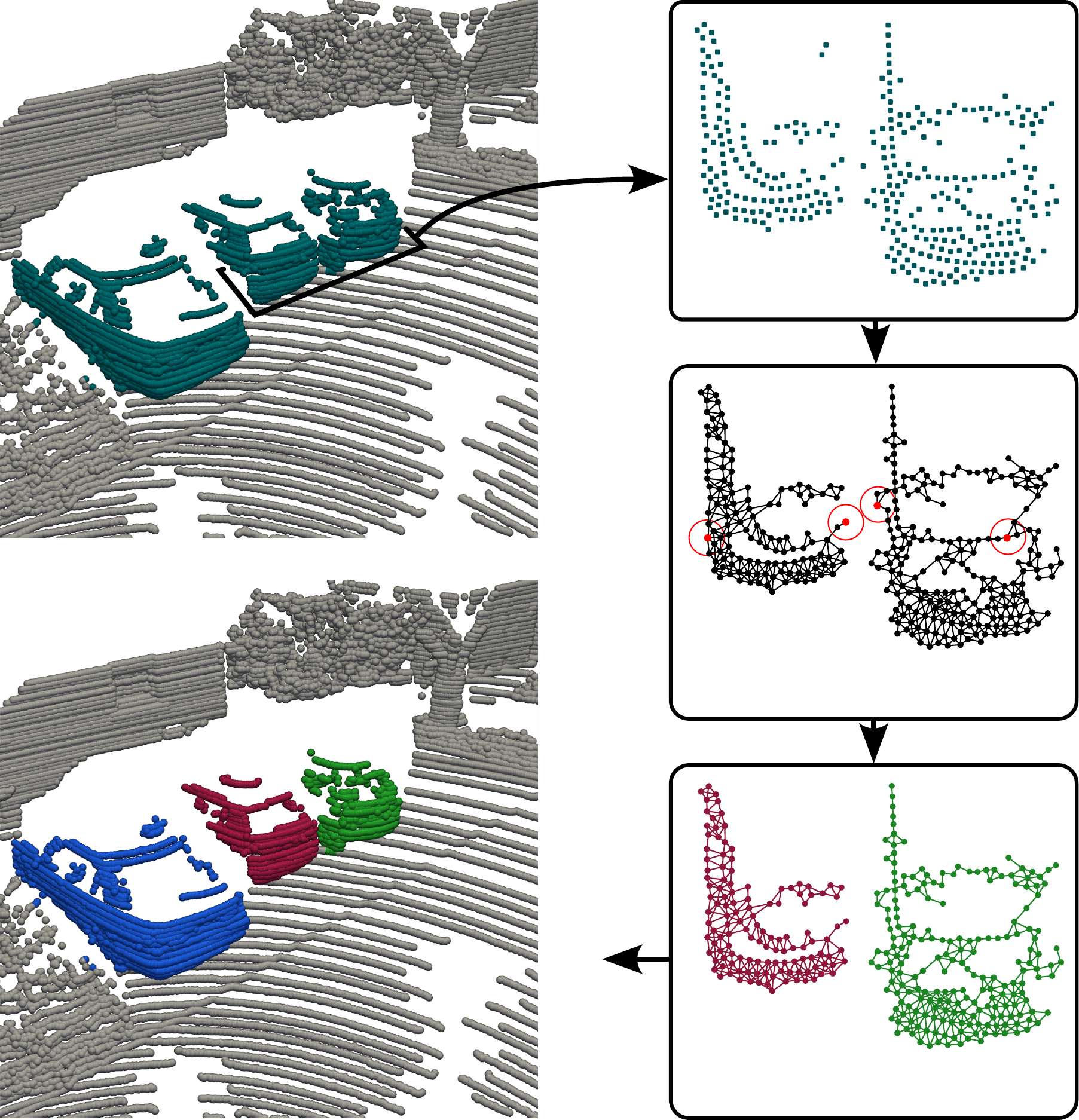}}%
    
    \put(0.08,1.04){\color[rgb]{0,0,0}\makebox(0,0)[t]{\lineheight{6.}\smash{\begin{tabular}[t]{c}\\Semantic\\predictions\end{tabular}}}}%

    \put(0.08,0.47){\color[rgb]{0,0,0}\makebox(0,0)[t]{\lineheight{6.}\smash{\begin{tabular}[t]{c}Instances\\predictions\end{tabular}}}}%

    \put(0.63823696,0.757){\color[rgb]{0,0,0}\makebox(0,0)[lt]{\lineheight{6.}\smash{\begin{tabular}[t]{l}BEV projection\end{tabular}}}}%
    
    \put(0.63823696,0.413){\color[rgb]{0,0,0}\makebox(0,0)[lt]{\lineheight{6.}\smash{\begin{tabular}[t]{l}Radius-based\\graph construction\end{tabular}}}}%
    
    \put(0.63823696,0.045){\color[rgb]{0,0,0}\makebox(0,0)[lt]{\lineheight{6.}\smash{\begin{tabular}[t]{l}Connected\\component extraction\end{tabular}}}}%

  \end{picture}%
\endgroup%

    \caption{\textbf{\ours clustering.} For a given semantic class a) we project the points in the BEV space (subsampled on the figure for visualization purpose), b) we build a kNN graph and filter by edge length and c) we extract the connected components.}
    \label{fig:clustering}
\end{figure}
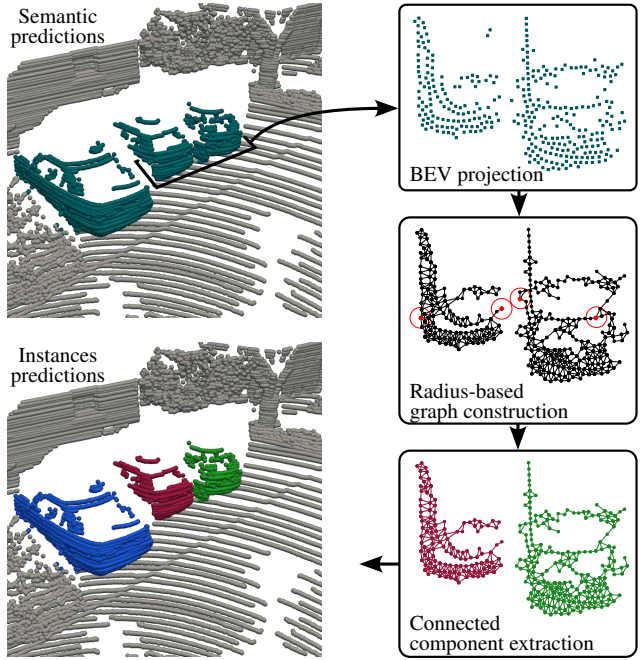

To move autonomously
in an outdoor environment, an agent must understand and segment its surroundings into categories. With a LiDAR, it involves recognizing the semantics of points and identifying individual instances of ``things'' (e.g., 'cars' or 'pedestrians').
This capability, known as LiDAR panoptic segmentation, is crucial for tasks such as object avoidance, trajectory forecasting, and path planning.

State-of-the-art \gilles{(SOTA)} approaches typically combine semantic segmentation and instance segmentation by training end-to-end networks using query-based mechanisms~\cite{maskpls,p3former,dqformer,ieqlps} or regressing instance centers~\cite{phnet, cfnet, centerlps}. Their architectures usually involve instance prediction heads, which require panoptic labels for training.
Instead, early methods~\cite{rbnn,supervoxels,clustering_1,slr,dividemerge} proposed unsupervised, learning-free, clustering algorithm to extract instances, sometimes completed by semantic segmentation networks to obtain panoptic segmentation~\cite{dividemerge}.
These methods appear to perform poorly in panoptic segmentation benchmark. One is easily tempted to disregard them and prefer to use end-to-end panoptic segmentation networks requiring costly panoptic segmentation labels.

This paper demonstrates that achieving \gilles{SOTA} panoptic segmentation does not require end-to-end networks or panoptic labels. The early pipeline combining semantic segmentation and unsupervised clustering suffices, as validated by extensive comparisons with current LiDAR panoptic methods.
We achieve this result thanks to the use of modern, high-performing semantic segmentation networks and to a clustering algorithm inspired by~\cite{less} with key changes to boost its performance including working per semantic class and in Bird's eye view rather than 3D.

This work has several practical consequences:
(a)~Thanks to the performance of modern semantic segmentation networks, it is possible to compete with SOTA methods without using instance labels.
(b)~The architecture is simple since it does not require a trainable instance head.
(c)~The method can be directly applied on top of any semantic prediction, without adaptation or access to underlying features.
(d)~As the clustering algorithm is extremely fast, the panoptic segmentation can be produced at high frequency, as required for embedding
on an autonomous vehicle, while using very little compute resources.

\begin{figure*}[t]
    \centering

    \setlength{\tabcolsep}{3pt}
    \begin{tabular}{@{}c|c|c@{}}
    \includegraphics[trim={0cm 4cm 0cm 4cm},clip,width=0.32\linewidth]{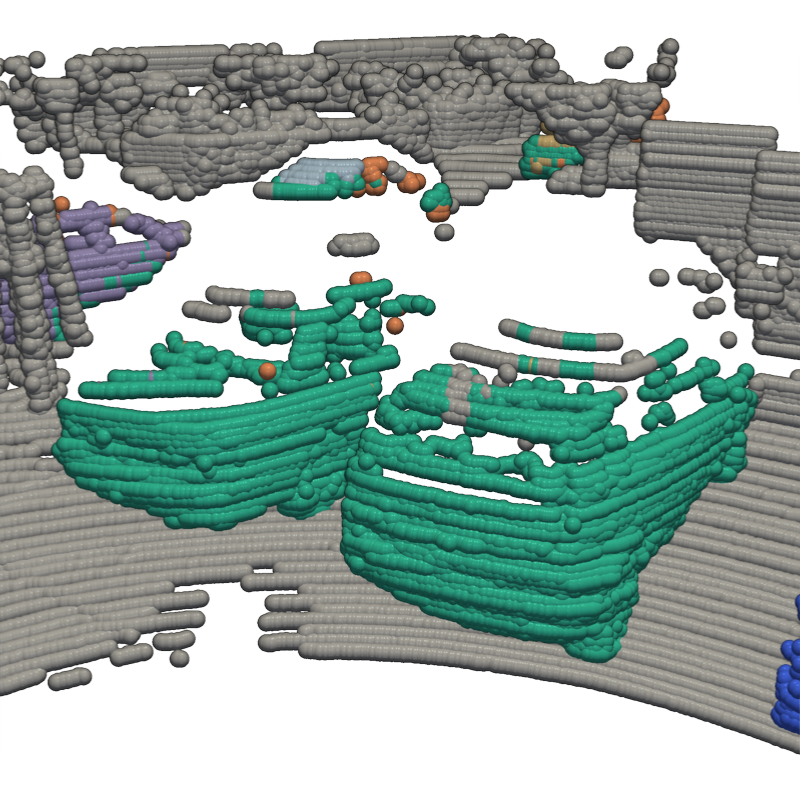}
    &
    \includegraphics[trim={0cm 4cm 0cm 4cm},clip,width=0.32\linewidth]{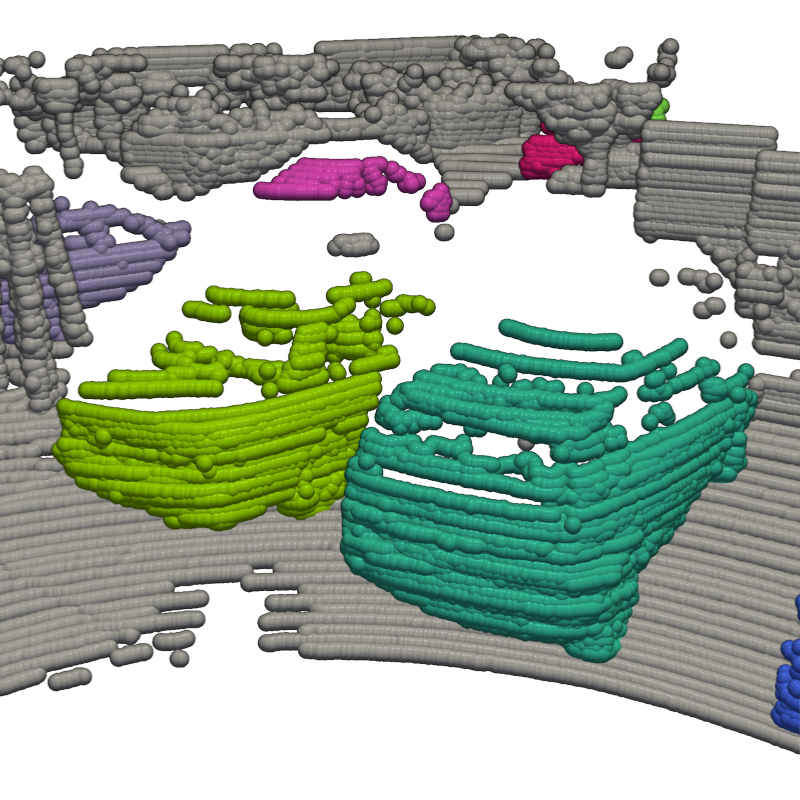}
    &
    \includegraphics[trim={0cm 4cm 0cm 4cm},clip,width=0.32\linewidth]{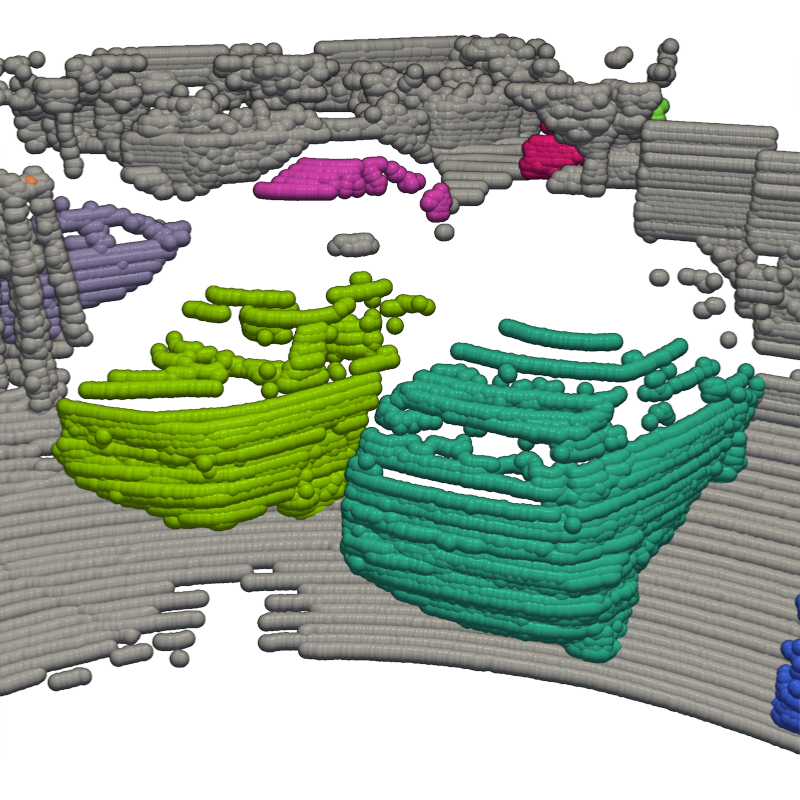}
    \\
    D\&M~\cite{dividemerge} w.~MinkUNet & \ours w.~MinkUNet & Ground Truth
    \end{tabular}

    \caption{\textbf{Examples of Instance predictions on SemanticKITTI.} We present the results obtained with D\&M~\cite{dividemerge} and \ours restricted to the \emph{car} class. Both methods are training-free clusterings, and use the same MinkUNet to obtain pointwise semantic predictions. When compared to the Ground Truth, we notice that D\&M does not satisfactorily separate the cars while \ours segment them correctly.}
    \label{fig:qualitative_example}
\end{figure*}

This clustering, which we call \ours for ``A Light Panoptic INstance Extractor'', works by constructing a nearest-neighbor graph on semantically consistent points, followed by some refinement using a box splitting mechanism. 
\ours builds on top of any off-the-shelf semantic segmentation network.
It is a learning-free method that does not require any knowledge of instance labels relying instead on a rough estimate of per-class object size, which is set using publicly available information found on the web.
This makes \ours a tuning free method, easy to use on any LiDAR dataset given a semantic segmentation backbone.
ALPINE is thus a strong baseline that can be used as reference before training with any instance labels in order to measure the benefits of such additional annotations.
\section{Related Work}
\label{sec:related_work}

\begin{figure*}
    \centering
    \includegraphics[width=\linewidth]{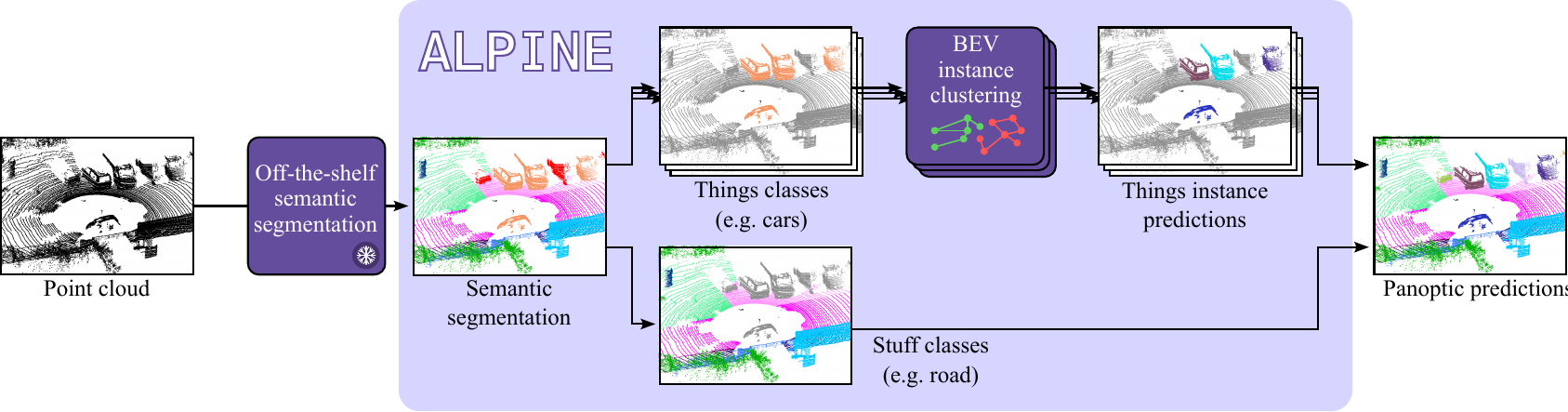}
    \caption{\textbf{Overview.} In \ours we take the output of a semantic segmentation model and apply our clustering algorithm on each \textit{things} classes to obtain instance masks and form panoptic predictions.}
    \label{fig:pipeline}
\end{figure*}

\subsection{Semantic segmentation} 

Semantic segmentation permits us to separate, at the point level, the main types of objects in a scene. Several methods and backbones have been designed for LiDAR semantic segmentation. They are usually classified under four main categories: point-based~\cite{pointnet,pointnetpp,pointnext,pointmixer,waffleiron} with the most recent ones leveraging a transformer architecture \cite{pointtransformer,ptv3,ptv2}, projection-based~\cite{rangenetpp,SalsaNet,polarnet,rangeviT,rangeformer,SqueezeSegV3}, sparse convolution-based~\cite{minkunet,cylinder3d,sdseg3d,af2s3net}, fusion-based when leveraging different point cloud representations~\cite{tornadonet,spvcnnpp,rpvnet,gfnet}.
 
We present a method that requires semantic predictions as input, regardless of how they are obtained. We test our method using different architectures for semantic segmentation: MinkUNet~\cite{minkunet}, WaffleIron~\cite{waffleiron}, PTv3~\cite{ptv3}.
We chose these methods for their competitive performance and diverse architectures: voxel-based using sparse convolutions, point-based using successive projections on 2D planes, point-based using attention layers.

\subsection{Panoptic segmentation}

Panoptic segmentation is an extension of semantic segmentation where one should also segments all instances of objects in ``thing'' classes.

\paragraph{Detection-based methods.} The first methods for panoptic segmentation were built on top of networks for semantic segmentation and object detection. The object detection branch puts boxes around each object instance, and the semantic segmentation branch is used to obtain the semantic class of each point.
Baseline methods are constructed using off-the-shelf detector~\cite{pointpillars,pointrcnn,second} and segmentor~\cite{kpconv,spvcnnpp,minkunet,af2s3net}.
More efficient methods often train the two tasks jointly ~\cite{efficientlps,mopt,aopnet,lidarmultinet,lpst}, with additional losses specifically designed for panoptic segmentation.

\paragraph{Clustering-based methods.} These type of methods also appeared quite early. Our method actually falls in the category.
These methods works in two successive steps~\cite{technicalsurvey}. 
A semantic segmentation backbone is first used to assign a class to each point. The points labeled as ``thing'' are then clustered to find instances.

One of the earliest method, the Euclidean Cluster~\cite{technicalsurvey,pcl,rusu2010semantic} consists in sequentially visiting each point, assigning to this point a new instance index if it is not already labeled, and labeling all its unlabeled neighbors within a certain radius with the same instance index. 
Another proposition consists in adapting  SLIC~\cite{slic}  from images to point clouds, using voxel representations~\cite{supervoxels,technicalsurvey}.
More advanced methods~\cite{slr,dividemerge} design a clustering method tailored for LiDAR point cloud, by taking into consideration the scan-line nature of the sensor's capture.
The hyperparameters of these methods are typically tuned according to expected object sizes and/or sensor properties.

More recent methods ~\cite{dsnet,dsnet2,panopticpolarnet,phnet,lidarpanoptic,smacseg,panoster,cfnet,panet,lcps,locationguided,pvcl,scan,lcps} leverage panoptic annotations and replace some hand-crafted steps of the clustering process by learnable steps.
For example, GP-S3Net~\cite{gps3net} creates a graph for points of thing classes using learned point features to obtain edge features, then perform edge classification.

Among clustering-based methods, a common approach consists in regressing for each point the position of the corresponding instance centroid, allowing an easier clustering done on the predicted cluster centers~\cite{dsnet,dsnet2,cpseg,centerlps,cfnet}, while having an easy to define training objective.
This approach is refined in Panoptic-PHNet where centroid-offset prediction and clustering are performed in the Bird's-Eye-View (BEV) to mitigate vertical over-segmentation. 

\paragraph{Query-based methods.}
These methods~\cite{maskrange,maskpls,dqformer,p3former,ieqlps} build upon MaskFormer~\cite{maskformer}, an image segmentation method.
To propose an end-to-end unified framework for panoptic segmentation, they create a set of (usually learnable) queries fed into a transformer decoder in charge of (softly) assigning points and queries. Each query encodes a different instance. Points responding maximally to the same query are part of the same instance.
In these architectures, the point features given to the transformer decoder can be obtained from different backbones working on range-view~\cite{maskrange}, voxels~\cite{maskpls}, BEV~\cite{dqformer}, or using both cartesian and polar space~\cite{p3former} representation of the point cloud.

\paragraph{Use of annotations.} Some datasets offer annotations for multiple tasks such as semantic \& panoptic segmentation and object detection~\cite{nuscenes}. Some methods leverage all the available annotation and train a single model on all tasks jointly, benefiting from the additional supervision~\cite{lidarmultinet,polarstream}. At the other end of the spectrum, some methods~\cite{3duis,tarl,unit} segment objects without using any annotation. However, these method are unable to predict semantics. 
In this work, we take inspiration from early clustering-based methods using only semantics annotations.
\section{{A strong baseline for panoptic segmentation}}
\label{sec:method}

Our approach is designed around three main components: a fast clustering algorithm having its roots in RBNN~\cite{rbnn} and LESS~\cite{less}, an annotation-free and training-free parameter selection process, and a refinement mechanism based on box splitting.

\subsection{\ours clustering}
\label{sec:clustering}

The overview of our instance clustering process is depicted in \cref{fig:clustering}. It is inspired by the clustering done in \cite{less} and works in four main steps.

\paragraph{One semantic class at a time.} After semantic prediction, we consider each ``thing'' class $c$ independently. 
We denote by $P_c$ the original point cloud $P$ restricted to the points with predicted label $c$. We denote $n_c$ the size of $P_c$. Each point cloud $P_c$ is processed separately.

\paragraph{Projection in BEV.} Inspired by previous works \cite{pointpillars,bevcontrast,phnet} leveraging the fact that objects of ``thing'' classes are rarely stacked vertically, we project the 3D points in the Bird's Eye View (BEV) representation (orthogonal projection onto the $(x, y)$ plane). In the following, we assume that points clouds $P_c$ are therefore two-dimensional.

\paragraph{$k$-nearest neighbors graph.} For each point in $P_c$, we obtain the list of its $k$ closest neighbors (typically $k=32$), using a 2D-Tree~\cite{kdtree}. The distances between the points are computed using only the point coordinates in the BEV representation, and therefore do not involve any learned features. We then construct a directed graph $\mathcal{G}$ where the nodes are the points in $P_c$ and each point is connected by an edge to its $k$ nearest neighbors, at most. We remove connections between points if the distance in BEV is larger than a threshold (see \cref{sec:distance_threshold}). Finally, the adjacency matrix of this directed graph is made symmetric by adding missing half-edges to obtain an undirected graph.
The choice of $k$ is not very critical for the method; it needs to be large enough not to miss too many connections with isolated points, while keeping it small enough to benefit from fast runtime. This is verified experimentally in the supplementary material.
The complexity of the construction of the 2D-Tree is $\mathcal{O}(n_c\log n_c)$, while the search for the $k$ nearest neighbors for all points is $\mathcal{O}(n_c\,k \log n_c)$. 

\paragraph{Connected components.} The clusters themselves are obtained by identifying the connected components in the constructed graph. The complexity of this algorithm is $\mathcal{O}(n_c k)$ in the worst case. 
We associate to each component a different instance index.

\begin{figure}[t]
    \centering

    \setlength{\tabcolsep}{3pt}
    \begin{tabular}{@{}c|c@{}}
    \includegraphics[trim={5cm 0 5cm 0},clip,width=0.48\linewidth]{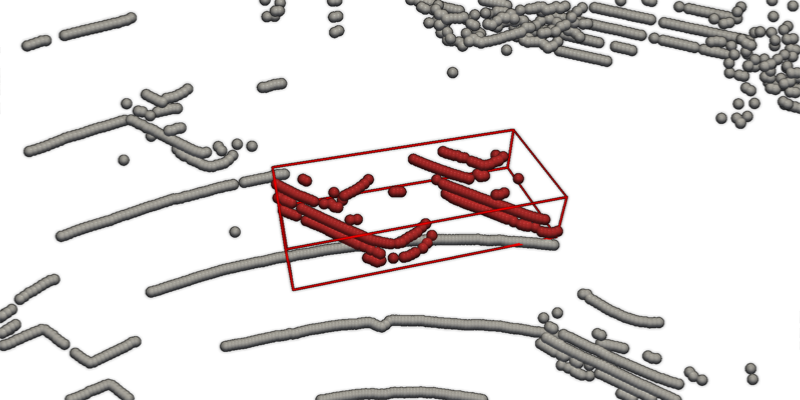}
    &
    \includegraphics[trim={5cm 0 5cm 0},clip,width=0.48\linewidth]{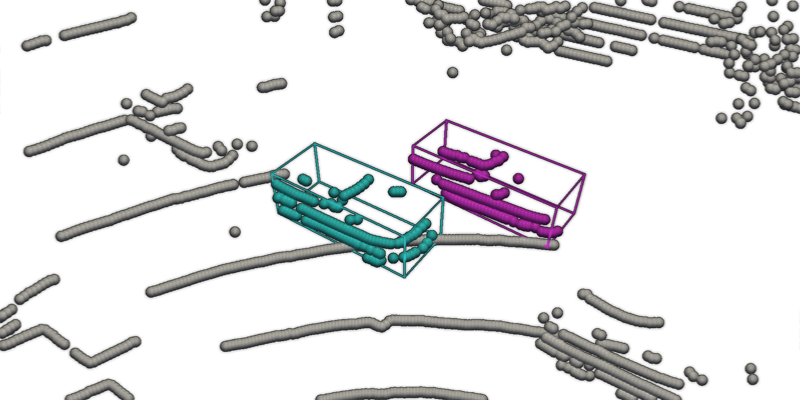}
    \\
    \midrule
    \includegraphics[trim={2cm 3cm 2cm 1cm},clip,width=0.48\linewidth]{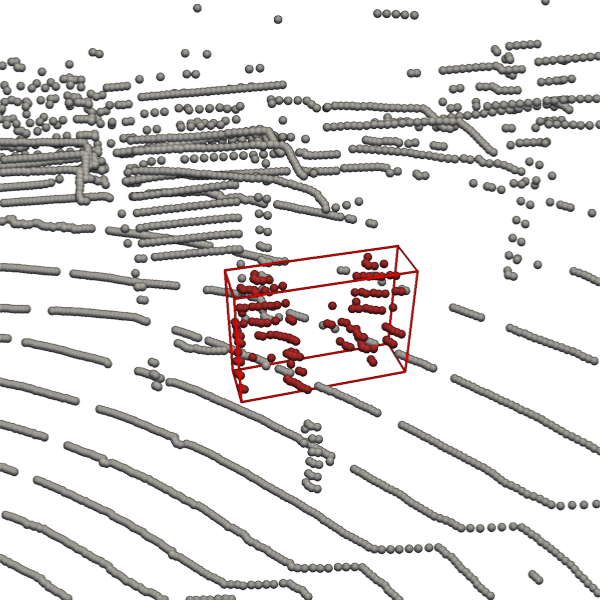}
    &
    \includegraphics[trim={2cm 3cm 2cm 1cm},clip,width=0.48\linewidth]{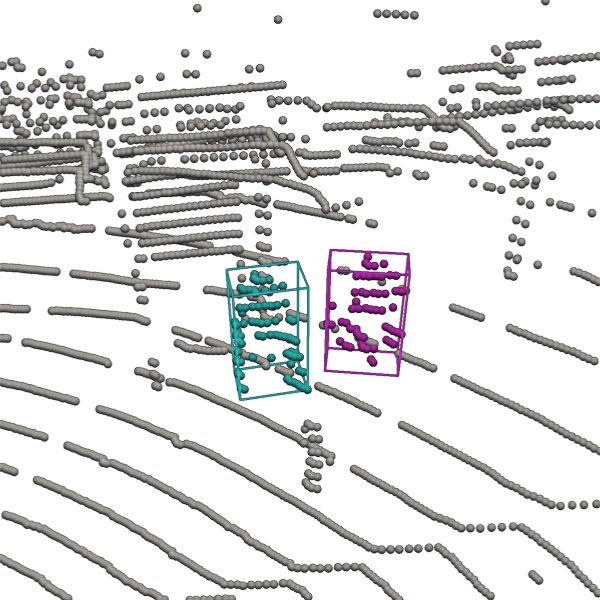}
    \end{tabular}

    \caption{\textbf{Example of bounding box splittings.} In the top examples, two cars are parked close to each other. While merged by the clustering, they do not fit in a car's box and the cluster is then split. In the bottom examples, the same mechanism is applied to two pedestrians in a bus shelter. Boxes are shown in 3D for illustrative purposes but the mechanism is purely 2-dimensional.}
    \label{fig:box_splitting}
\end{figure}

\paragraph{Key differences with LESS and RBNN.} \ours builds upon the strong graph-based clustering foundations laid by RBNN~\cite{rbnn} and LESS~\cite{less}; however applying a few key changes making it both stronger and more practical to use. (1) Unlike in both other methods, the adjacency of each point is limited to at most $k$ reducing the complexity of the algorithm from $\mathcal{O}(n_c^2)$ to $\mathcal{O}(n_c\,k \log n_c)$, scaling better to large clusters; (2) it is also applied class-wise, (3) and in BEV, with a per-class constant threshold. Besides, while conceptually similar, RBNN did not use connected components to find clusters, relying instead on a custom cluster merging scheme. Contrary to ours, LESS' clustering was proposed as a pre-annotation tool only.

\subsection{Choice of distances and thresholds} \label{sec:distance_threshold}

In order to decide whether two points should be connected in $\mathcal{G}$, and thus should belong to the same instance, we use a threshold $t_c$ on the distance between two points in BEV.

\paragraph{Class-wise thresholds.}
When considering whether two points should be connected, an important quantity to consider is the typical size of the objects to segment: points that are $3$ meters apart are never going to belong to the same pedestrian, but might belong to the same truck. We therefore use separate thresholds $t_c$ for each semantic class $c$. The only information that we require to set $t_c$ is the an estimate of the average size of the objects in each class. 

This information can be obtained \emph{without any ground-truth annotation, and even without any access to the dataset}. 
For example, for cars, one can retrieve from the Internet the average size of vehicles where the system is to be deployed.
For each class, we can thus get an estimate of the typical size of bounding boxes around objects in this class. We set $t_c$ to the smallest side of this reference bounding box.
More considerations as to how we obtained such approximate average boxes are given in \cref{sec:parameters}.

This parametrization addresses a known domain shift issue related to sizes when training and testing in different countries \cite{wang2020trainingermany}. It enables a simple form of domain adaptation where only a few scalars $t_c$ have to be set from one country to another, besides any adaptation or generalization regarding the underlying semantics.

\paragraph{Range vs threshold.}
The distance between two points in a LiDAR point clouds naturally increases with their range (distance between a point and the sensor): LiDAR point clouds are less dense far away from the sensor. For this reason, to similarly create object clusters and decide if an edge between two points should be removed or not, LESS~\cite{less} used a threshold that varies with the range. We tested such a strategy in \cref{sec:ablation} but found no improvement in doing so. We thus simply use a constant class-wise threshold.

\subsection{Box splitting}
\label{sec:box_splitting}

To refine our segmentation further, we use a box-splitting strategy.
When a cluster of a given semantic class $c$ does not fit within the reference bounding box of this class (as defined in \cref{sec:distance_threshold}),
enlarged by a constant (not class-wise) proportional margin (of 30\%) to account for bigger-than-average objects, it likely actually contains multiple instances of this class.
In this case, our box-splitting method performs a binary search on the threshold parameter $t$ to recursively find the largest clusters that all fit into the reference bounding box. The pseudo-code for this algorithm is given in \cref{alg:split_cluster}, and an illustrative example is provided in \cref{fig:box_splitting}. The box-fitting procedure that is used internally in the algorithm is borrowed from the literature~\cite{modest} and discussed in the appendix.

\SetInd{0.25em}{0.5em} 
\begin{algorithm}
  \caption{\textbf{Box splitting algorithm.} \small This algorithm splits clusters that do not fit in $B$ into sub-clusters that all fit in $B$. \texttt{\bfseries{box\_fitting}} returns the smallest bounding box encompassing $P$ and \texttt{\bfseries{clusterize}} is the clustering algorithm described in \cref{sec:clustering}.}
  \label{alg:split_cluster}
  \small
  \Function{split\_cluster(P, B, t)}{
  \Input{Points \var{P}, Average box \var{B}, Threshold \var{t}}
  \Output{List of cluster points}
  \If{$\FuncCall{box\_fitting}{P} \text{\rm fits in } \var{B}$\:}{
  \Return{[\var{P}]} \Comment{Points already fit in \var{B}}
  }
  \Else{
   $\var{t} \assign \var{t} / 2$\;
   $\var{dt} \assign \var{t}$\;
   \While{\:\KwSty{true}\:}{
   $\var{dt} \assign \var{dt} / 2$\;
   $\var{C} \assign \FuncCall{clusterize}{P, t}$\;
   \If{\FuncCall{len}{C} = 1\:}{
   $\var{t} \assign \var{t} - \var{dt}$\Comment{Decrease threshold}
   }
   \ElseIf{\FuncCall{len}{C} > 2\:}{
   $\var{t} \assign \var{t} + \var{dt}$\Comment{Increase threshold}
   }
   \Else{
   \Return{\FuncCall{split\_cluster}{C[0], B, t} + \FuncCall{split\_cluster}{C[1], B, t}}
   }
   }
  }
  }
  \normalsize
\end{algorithm}

\section{Experiments}
\label{sec:expe}

\begin{table}[ht]
    \small 
    \centering
    \setlength{\tabcolsep}{2.6pt}
    \begin{NiceTabular}{@{}l@{}l|@{\hskip 1.5pt}c@{\hskip 1.5pt}c@{\hskip 1.5pt}c@{\hskip 1.5pt}|cccc|c@{}}
    \toprule
    \multirow{2}{*}{Method} &&\footnotesize Inst.& \multirow{2}{*}{\footnotesize TTA} & \multirow{2}{*}{\footnotesize Ens.} & \multirow{2}{*}{PQ} & \multirow{2}{*}{PQ\textsuperscript{$\dagger$}} & \multirow{2}{*}{RQ} & \multirow{2}{*}{SQ} & \multirow{2}{*}{mIoU} \\
    && \footnotesize lbs. & &\\
    \midrule
    

    
    
    

    LCPS (lidar)               & \footnotesize \cite{lcps}               & \cmark & \xmark & \xmark &
    55.7 & 65.2 & 65.8 & 74.0 & 61.1 \\ 
    
    D\&M ({\scriptsize Cyl.3D})         & \footnotesize \cite{dividemerge}         & \xmark & \xmark & \xmark &
    57.2 & -    & -    & -    & - \\ 

    
    Panop.-PolarNet          & \footnotesize \cite{panopticpolarnet}    & \cmark & \xmark & \xmark &
    59.1 & 64.1 & 70.2 & 78.3 & 64.5 \\ 
    
    
    MaskPLS                    & \footnotesize \cite{maskpls}            & \cmark & \xmark & \xmark &
    59.8 & -    & 69.0 & 76.3 & 61.9 \\ 

    DS-Net ({\scriptsize SPVCNN})            & \footnotesize \cite{dsnet2}             & \cmark & \qmark & \qmark &
    61.4 & 65.2 & 72.7 & 79.0 & 69.6 \\ 
    
    Panop.-PHNet             & \footnotesize \cite{phnet}               & \cmark & \xmark & \xmark &
    61.7 & -    & -    & -    & 65.7 \\ 
    
    PANet                      & \footnotesize \cite{panet}               & \cmark & \qmark & \qmark &
    61.7 & 66.6 & 71.8 & 79.3 & 68.1 \\ 

    D\&M ({\scriptsize Mink.}) & \footnotesize \cite{dividemerge} & \xmark & \mmark & \xmark & 
    61.8 & 66.2 & 72.8 & 79.6 & 71.4\\

    CenterLPS                  & \footnotesize \cite{centerlps}         & \cmark & \qmark & \qmark & 
    62.1 & 67.0 & 72.0 & 80.7 & 68.1 \\ 
    
    \rowcolor{green!15}
    \ours ({\scriptsize PTv3})            &                          & \xmark & \mmark & \xmark & 
    62.4 & 66.2 & 72.0 & 76.7 & 67.3 \\

    P3Former                   & \footnotesize \cite{p3former}          & \cmark & \xmark & \xmark & 
    62.6 & 66.2 & 72.4 & 76.2 & -    \\
    
    CFNet                      & \footnotesize \cite{cfnet}        & \cmark & \xmark & \xmark & 
    62.7 & 67.5 & -    & -    & 67.4 \\ 
    
    LPST                       & \footnotesize \cite{lpst}                & \cmark & \xmark & \xmark & 
    63.1 & 70.8 & 73.1 & 79.2 & 69.7 \\ 
    
    GP-S3Net              & \footnotesize \cite{gps3net}           & \cmark & \qmark & \qmark & 
    63.3 & 71.5 & 75.9 & 81.4 & 73.0 \\ 
    
    DQFormer                   & \footnotesize \cite{dqformer}           & \cmark & \xmark & \xmark &
    63.5 & 67.2 & 73.1 & 81.7 & - \\ 

    \rowcolor{green!15}
    \ours ({\scriptsize WI})            &                          & \phantom{*}\xmark* & \mmark & \xmark & 
    64.2 & 69.0 & 74.1 & 79.7 & 70.3 \\

    \rowcolor{green!15}
    \ours ({\scriptsize Mink.})    &                          & \xmark & \xmark & \xmark & 
    64.2 & 68.9 & 74.1 & 84.4 & 70.7 \\
    
    PUPS (w/o ens.)            & \footnotesize \cite{pups}              & \cmark & \xmark & \xmark & 
    64.4 & 68.6 & 74.1 & 81.5 & - \\ 

    IEQLPS                     & \footnotesize \cite{ieqlps}            & \cmark & \qmark & \qmark & 
    64.7 & 68.1 & 74.7 & 81.3 & - \\ 
    
    \rowcolor{green!15}
    \ours ({\scriptsize Mink.})         &                          & \xmark & \mmark & \xmark & 
    65.9 & 70.2 & 75.5 & 81.4 & 72.2 \\
    
    PUPS              & \footnotesize \cite{pups}             & \cmark & \cmark & \cmark & 
    66.3 & 70.2 & 75.6 & 82.5 & - \\ 

    \multicolumn{2}{>{\columncolor{green!15}}l|}{\cellcolor{green!15}\hspace{-3pt}\ours ({\scriptsize Mink.+PTv3})} & \cellcolor{green!15}\xmark & \cellcolor{green!15}\mmark & \cellcolor{green!15}\mmark &  
    \cellcolor{green!15}66.6 & \cellcolor{green!15}70.8 & \cellcolor{green!15}76.1 & \cellcolor{green!15}82.6 & \cellcolor{green!15}72.0 \\
    \bottomrule
    \multicolumn{10}{l}{Mink.: MinkUNet~\cite{minkunet} \hspace{15pt} WI: WaffleIron-256~\cite{waffleiron}}\\
    \multicolumn{10}{l}{*: the publicly available model for WI-256~\cite{waffleiron} used instance} \\
    \multicolumn{10}{l}{~~~~annotations in its data augmentation pipeline}
    \end{NiceTabular}
    \caption{Panoptic segmentation results on the validation set of SemanticKITTI. `TTA' and `Ens.' stand for Test-Time Augmentation and Ensembling. \mmark{} denotes that TTA/ensemble was used on the semantic head only. The main metric is the PQ. A more detailed version of this table is available in the supplementary material.}
    \label{tab:results_semkitti_val_compressed}
\end{table}

\begin{table}
    \small
    \centering
    \setlength{\tabcolsep}{2.7pt}
    \begin{NiceTabular}{@{}l@{}l|@{\hskip 1.5pt}c@{\hskip 1.5pt}c@{\hskip 1.5pt}c@{\hskip 1.5pt}|cccc|c@{}}
    \toprule
    
    \multirow{2}{*}{Method} &&\footnotesize Inst.& \multirow{2}{*}{\footnotesize TTA} & \multirow{2}{*}{\footnotesize Ens.} & \multirow{2}{*}{PQ} & \multirow{2}{*}{PQ\textsuperscript{$\dagger$}} & \multirow{2}{*}{RQ} & \multirow{2}{*}{SQ} & \multirow{2}{*}{mIoU} \\
    && \footnotesize lbs. & &\\

    \midrule
    
    
    
    
    
    
    
    
    DS-Net({\scriptsize SPVCNN})   & \footnotesize \cite{dsnet2}                            & \cmark & \qmark & \qmark & 
    64.7 & 67.6 & 76.1 & 83.5 & 76.3 \\ 
    
    
    
    Panop.-PolarNet & \footnotesize \cite{panopticpolarnet}                 & \cmark & \xmark & \xmark & 
    67.7 & 71.0 & 78.1 & 86.0 & 69.3 \\ 
    
    
    PANet             & \footnotesize \cite{panet}                            & \cmark & \qmark & \qmark & 
    69.2 & 72.9 & 80.7 & 85.0 & 72.6 \\ 
    
    CPSeg             & \footnotesize \cite{cpseg}                            & \cmark & \qmark & \qmark & 
    71.1 & 75.6 & 82.5 & 85.5 & 73.2 \\ 
    
    LCPS (lidar)      & \footnotesize \cite{lcps}                             & \cmark & \xmark & \xmark & 
    72.9 & 77.6 & 82.0 & 88.4 & 75.1 \\ 
    
    PUPS              & \footnotesize \cite{pups}                             & \cmark & \qmark & \qmark & 
    74.7 & 77.3 & 83.3 & 89.4 & - \\
    
    Panop.-PHNet    & \footnotesize \cite{phnet}                            & \cmark & \xmark & \xmark & 
    74.7 & 77.7 & 84.2 & 88.2 & 79.7 \\ 
    
    CFNet             & \footnotesize \cite{cfnet}                            & \cmark & \xmark & \xmark & 
    75.1 & 78.0 & 84.6 & 88.8 & 79.3 \\
    
    P3Former          & \footnotesize \cite{p3former}                         & \cmark & \xmark & \xmark & 
    75.9 & 78.9 & 84.7 & 89.7 & - \\ 
    
    CenterLPS         & \footnotesize \cite{centerlps}                        & \cmark & \qmark & \qmark & 
    76.4 & 79.2 & 86.2 & 88.0 & 77.1\\ 

    \rowcolor{green!15}
    \ours ({\scriptsize WI})  & & \xmark &  \xmark & \xmark & 
    76.9 & 79.9 & 85.7 & 89.3 & 80.3 \\
    
    IEQLPS            & \footnotesize \cite{ieqlps}                           & \cmark & \qmark & \qmark & 
    77.1 & 79.1 & 86.5 & 88.2 & - \\ 

    LPST               & \footnotesize \cite{lpst}                            & \cmark & \xmark & \xmark & 
    77.1 & 79.9 & 86.5 & 88.6 & 80.3 \\ 
    
    DQFormer          & \footnotesize \cite{dqformer}                         & \cmark & \xmark & \xmark & 
    77.7 & 79.5 & 86.8 & 89.2 & - \\ 
    
    \rowcolor{green!15}
    \ours ({\scriptsize WI})  & & \xmark & \mmark & \xmark &
    77.9 & 80.7 & 86.5 & 89.6 & 81.4\\
    
    \rowcolor{green!15}
    \ours ({\scriptsize PTv3})    & & \xmark & \mmark & \xmark & 
    78.9 & 81.3 & 87.0 & 90.4 & 81.5\\

    \multicolumn{2}{>{\columncolor{green!15}}l|}{\cellcolor{green!15}\hspace{-3pt}\ours ({\scriptsize WI+PTv3})}& \cellcolor{green!15}\xmark & \cellcolor{green!15}\mmark & \cellcolor{green!15}\mmark & 
    \cellcolor{green!15}79.5 & \cellcolor{green!15}81.9 & \cellcolor{green!15}87.6 & \cellcolor{green!15}90.5 & \cellcolor{green!15}82.7 \\

    LidarMultiNet     & \footnotesize \cite{lidarmultinet}                    & \cmark & \cmark & \cmark & 
    81.8 & -    & 89.7 & 90.8 & 83.6 \\ 

    \bottomrule
    \multicolumn{10}{l}{WI: WaffleIron-768~\cite{waffleiron}}\\
    \end{NiceTabular}
    \caption{Panoptic segmentation results on the validation set of nuScenes. `TTA' and `Ens.' stand for Test-Time Augmentation and Ensembling. \mmark{} denotes that TTA/ensemble was used on the semantic head only. The main metric is the PQ. A more detailed version of this table is available in the supplementary material.}
    \label{tab:results_nuscenes_val_compressed}
\end{table}

\begin{table}[t]

    \small
    \centering
    \setlength{\tabcolsep}{2.5pt}
    
    \begin{NiceTabular}{l@{}l|c|cccc|c}

    \toprule
    \multirow{2}{*}{Method} &&\footnotesize Instance & \multirow{2}{*}{PQ} & \multirow{2}{*}{PQ\textsuperscript{$\dagger$}} & \multirow{2}{*}{RQ} & \multirow{2}{*}{SQ} & \multirow{2}{*}{mIoU} \\
    && \footnotesize labels & &\\
    \midrule
    LPSAD (in \cite{gps3net}) & \footnotesize \cite{lidarpanoptic}& \cmark & 
    22.5 & 32.7 & 34.0 & 53.5 & 35.5\\ 
    TORNADO-Net                  & \footnotesize \cite{tornadonet}   & \cmark &
    33.7 & 43.3 & 46.0 & 68.4 & 44.5 \\ 
    DS-Net                                & \footnotesize \cite{dsnet}        & \cmark & 
    35.6 & 45.9 & 49.2 & 68.6 & 54.5 \\ 
    GP-S3Net                              & \footnotesize \cite{gps3net}      & \cmark & 
    48.7 & 60.3 & 63.7 & 61.3 & 61.8 \\ 
    \rowcolor{green!15}
    \ours ({\scriptsize PTv3})  &              & \xmark & 
    51.4 & 57.7 & 67.2 & 74.3 & 58.3 \\ 

    \bottomrule
    
    \end{NiceTabular}
    \caption{Panoptic segmentation results on Sequence $02$ of SemanticPOSS as validation set. The main metric is the PQ.}
    \label{tab:results_semposs_scene2_compressed}
\end{table}

\subsection{Datasets and metrics}

\textbf{nuScenes}~\cite{nuscenes} is a dataset captured in Boston and Singapore, with a $32$-beam LiDAR sensor.
It contains $10$ things classes and $6$ stuff classes. The instance labels were automatically generated by combining object detection and semantic segmentation labels, and were not manually curated. The results are reported on the official validation set.

\textbf{SemanticKITTI}~\cite{semantickitti} is a dataset captured around Karlsruhe in Germany with a 64-beam LiDAR sensor. It contains $8$ things classes and $11$ stuff classes. The instance labels were manually annotated. The results are reported on Sequence 08, commonly used as the validation sequence.

\textbf{SemanticPOSS}~\cite{semanticposs} is a smaller dataset captured in the surroundings of the Peking University with a $40$-beam LiDAR, focusing on having many dynamic (thing) objects. It contains only $3$ things classes, namely \relax{person}, \relax{rider} and \relax{car}, as well as $10$ stuff classes.
The results are reported on Sequence 02, used as a validation sequence as done in~\cite{gps3net}.

\vspace{4pt}

We use standard metrics, which are defined in Appendix.

\subsection{Size parameters setting}
\label{sec:parameters}

The parameters needed in \ours, as described in \cref{sec:distance_threshold,sec:box_splitting}, are set after estimating the average size of object bounding boxes in each class.

We consider two kinds of size parameter settings: the web-based setting obtains sizes from a few queries on the Internet; the dataset-based setting exploits ground-truth knowledge coming with datasets.

For web-based box sizes on nuScenes, we used the publicly available average size of cars in the United States in 2018 (15.6\,$\times$\,6.3\,ft)~\cite{carsize_us}; we made the simplifying assumption that buses, trailers, trucks and construction vehicles have size 10\,$\times$\,3\,m; we used standard bike and motorcycle sizes found online \cite{bikesize,motorcyclesize} ; we made the the approximation that pedestrians fit in box of half their arm span, estimates using~\cite{armspan,height}; and we made an educated size guess of 2\,$\times$\,0.5\,m for barriers and 40\,cm for cones.

Similar web-based information was exploited to set the sizes for SemanticKITTI. We used the average car size in Europe of 4.4\,$\times$\,1.8\,m~\cite{carsize_EUR}. The same found sizes for bicycle and motorcycle as in nuScenes~\cite{bikesize,motorcyclesize} as well as for bicyclist and motorcyclist.
The class person was treated as in nuScenes. For truck and other-vehicle, we made the same assumption as in nuScenes of a 10\,$\times$\,3\,m size.

For the $3$ classes of SemanticPOSS, we used the value we obtained for the equivalent class in SemanticKITTI, i.e. pedestrians, bicyclists and cars.

For dataset-based box sizes, we use the average size of the bounding boxes used in~\cite{openpcdet,second}. On nuScenes, these average size is provided for all $10$ thing classes. On SemanticKITTI, these sizes are only available for cars, pedestrians and bicyclists; we use the value computed on nuScenes for the 5 remaining classes. 
On SemanticPOSS, we just re-use the sizes obtained on SemanticKITTI.

In the end, we did not witness a significant change of performance between both sets of parameters in the experiments that follow and use by default, in all results, the annotation-free web-based sizes.

\subsection{Comparison to the state-of-the-art}

\begin{table}[t]
    \small
    \centering
    \setlength{\tabcolsep}{3pt}
    \begin{tabular}{l|cccccc}
    
     Rank/ Method & PQ & PQ\textsuperscript{$\dagger$} & RQ & SQ & PQ\textsubscript{Th} & mIoU \\
    
    \midrule
    5/ P3Former            & 65.3 & 67.8 & 74.9 & 86.6	& 67.4 & 66.1 \\
    4/ PUPS & 65.7 & 70.3 & 75.8 & 85.7 & 68.1 & 68.5 \\
    \rowcolor{green!15}
    3/ P3Former \& \ours   & 66.1 & 68.7 & 75.9 & 86.7 & 69.4 & 66.1\\
    2/ UniSeg & 67.2 & 72.1 & 78.1 & 85.5 & 67.5 & 73.8* \\
    \rowcolor{green!15}
    1/ UniSeg \& \ours & \textbf{70.2} & \textbf{75.1} & \textbf{80.5} & \textbf{86.7} & \textbf{74.3} & \textbf{75.1}* \\
    
    \end{tabular}
    \caption{\textbf{SemanticKITTI test set results} (official leaderboard, June 2025). We used P3Former's and UniSeg's semantic predictions and used \ours for instance segmentation. *(See \cref{sec:study_components})
    }
    \label{tab:results_semkitti_test}
\end{table}

\begin{table}[t]
    \small
    \centering
    \setlength{\tabcolsep}{3pt}    
\resizebox{\columnwidth}{!}{%
    \begin{tabular}{l|c|ccc|ccc|c}
    
    & sem. & \multicolumn{3}{c|}{Original head} &  \multicolumn{3}{c|}{\ours} & FPS \\ 
    
     Panop. seg. & mIoU & PQ & PQ\textsuperscript{$\dagger$} & PQ\textsubscript{Th} & PQ & PQ\textsuperscript{$\dagger$} & PQ\textsubscript{Th} & (Hz)
    \\
    
    \midrule
    DS-Net & 63.5 & 57.7 & 63.4 & 61.8 & \textbf{58.5} & \textbf{64.1} & \textbf{63.6} & 0.7
    \\
    Panoptic-P.Net & 64.4 & 58.9 & 63.9 & 65.2 & \textbf{59.5} & \textbf{64.5} & \textbf{66.7} & 1.6
    \\
    MaskPLS & 61.9 & 59.8 & 63.6 & 63.7 & \textbf{60.2} & \textbf{64.0} & \textbf{64.7} & N/A
    \\
    LCPS (lidar) & 64.3 & 61.3 & \textbf{65.6} & 70.0 & \textbf{61.4} & \textbf{65.6} & \textbf{70.1} & 1.8
    \\
    PANet & 67.9 & 61.2 & 66.1 & 66.7 & \textbf{61.3} & \textbf{66.3} & \textbf{66.9} & 7.1
    \\
    P3Former & 66.8 & 62.6 & 66.2 & 69.4 & \textbf{62.9} & \textbf{66.5} & \textbf{70.1} & N/A
    \\
    \end{tabular}}
    \vspace{-0.2cm}
    \caption{\textbf{Comparison between \ours and the instance segmentation capabilities of other methods}, on the validation set of Semantic\-KITTI. For each panoptic segmentation method with available code and checkpoint, we isolated the plain semantic predictions and report the mIoU~(`sem.'), and constructed on top of them our 
    instances (`\ours'), which we compare to the instances obtained by the original methods (`Original head'). An indicative instance prediction frequency (FPS) 
    is given after subtracting the time of the semantic forward pass to the total runtime.}
    \label{tab:comparisons}
\end{table}

On all datasets, we apply \ours\ combined with different backbones trained for semantic segmentation. We re-used the provided checkpoints when publicly available or retrain them ourselves with the provided code.

\paragraph{SemanticKITTI.} \cref{tab:results_semkitti_val_compressed} presents our results on the validation set of SemanticKITTI. On this dataset, we consider 3 different semantic segmentation backbones: PTv3~\cite{ptv3}, WI-256~\cite{waffleiron}, and MinkUnet~\cite{minkunet} from OpenPCSeg~\cite{openpcseg}. 

\ours is able to reach SOTA performance (PQ), outperforming, e.g., the end-to-end method PUPS \cite{pups}.
Moreover, when comparing to other methods at equivalent semantic segmentation mIoU, \ours is comparable or outperforms these methods: \ours w.~PTv3 vs CFNet \cite{cfnet}, PANet~\cite{panet}, CenterLPS~\cite{centerlps}; or \ours w.~MinkUNet vs GP-S3Net~\cite{gps3net}. It shows that \ours benefits easily from better semantic segmentation backbones.

\paragraph{nuScenes.} \cref{tab:results_nuscenes_val_compressed} presents our results on the validation set of nuScenes.
On this dataset, we consider two different semantic segmentation backbones: PTv3~\cite{ptv3} and WI-768~\cite{waffleiron}. 
We are only outperformed by LidarMultiNet~\cite{lidarmultinet}, which was trained using both semantics and object boxes, using ensembling and test-time augmentations (TTA) on both tasks, and benefits from a better semantic prediction. 

\paragraph{SemanticPOSS.} \cref{tab:results_semposs_scene2_compressed} presents results obtained on Sequence 02 of SemanticPOSS. We notice that \ours combined with PTv3~\cite{ptv3} reaches state-of-the-art results. 

\begin{table}[t]
    \centering
    \setlength{\tabcolsep}{3pt}
    \small
    \begin{tabular}{cc|cc|cc|cc}
    \toprule
    \multicolumn{2}{c}{\ours$\dagger$} & \multicolumn{2}{c}{DBSCAN} & \multicolumn{2}{c}{HDBSCAN} & \multicolumn{2}{c}{D\&M} \\
    PQ & FPS & PQ & FPS & PQ & FPS & PQ & FPS \\
    \midrule
    \textbf{65.5} &\textbf{14.4~Hz} & 56.7 & 3.2~Hz & 55.1 & 4.5~Hz & 61.8 & 0.5~Hz \\
    \bottomrule
    \end{tabular}
    \caption{
    \textbf{Comparing PQ and computation times of clustering methods} on the validation set of SemanticKITTI, applied to the same MinkUNet semantics. $\dagger$No box splitting.
    }
    \label{tab:clustering}
\end{table}

\subsection{Study of \ours's components}
\label{sec:study_components}

\paragraph{Comparison to other instance heads.}
We present results obtained with \ours on the test set of SemanticKITTI. First, we use the public model of P3Former for semantic prediction and extract object instances using \ours. The PQ which is at 65.3 with P3Former improves to 66.1 thanks to \ours, the third best entry on the leaderboard. Second, we repeat the same procedure with the semantic labels of UniSeg~\cite{uniseg} (labels kindly provided by the authors). UniSeg with \ours's instance labels ranks first on the leaderboard of SemanticKITTI. These results highlight the high quality of the instances extracted by our method. We note nevertheless that these results are not strictly obtained without instance labels as both UniSeg and P3Former did use them during training. 
We also remark that while the semantic mIoU of P3Former and P3Former\&\ours are identical, UniSeg\&ALPINE's semantic mIoU is above UniSeg's: 75.1 vs 73.8. We did not modify UniSeg's semantic labels (75.1 is in line with the score of UniSeg on the \emph{semantic} segmentation leaderboard). A misformatting in UniSeg's semantic labels reduces their score from 75.1 to 73.8, which we could verify on the semantic benchmark.

We conduct a more complete study on the validation set of SemanticKITTI.
\cref{tab:comparisons} shows how all panoptic methods for which we found code and checkpoints on SemanticKITTI compare against \ours\ \emph{at equal semantic prediction}: we use the semantic prediction of the original method but replace the instance prediction mechanism with \ours.
We show that \ours outperforms every single method, while using no annotation and requiring no GPU. 
This further indicates that current instance prediction heads do not perform as well as expected and can be outmatched by a simple clustering scheme without training nor instance annotation.
An indicative ``instance runtime'' of all methods have been obtained by subtracting the segmentation forward pass to the total runtime, when this was possible. This approximate measurement is comparable (while not directly equivalent) to the \ours FPS reported in \cref{tab:clustering}.

\paragraph{Comparison to other clustering methods.} \cref{tab:clustering} shows that \ours is both faster and better performing on our task than other clustering methods, namely the standard DBSCAN and HDBSCAN, as well as D\&M~\cite{dividemerge}, which was specifically designed for this task too. We are also within the limit of real-time processing, as SemanticKITTI is captured at 10\,Hz, while we use only limited computing resources (single-core CPU usage). In these experiments, all clustering methods use the same semantics, predicted by MinkUNet~\cite{minkunet}.
The box splitting was not used on any method in order to compare clustering capacity only. DBSCAN and HDBSCAN were performed on BEV, with fixed parameters of $\epsilon=1$ for DBSCAN and parameters gathered from 3DUIS~\cite{3duis} and $\epsilon$ proportional to $t_c$ for HDBSCAN; with the proportionality coefficient obtained with a fine parameter search. Those settings were those that we found worked best.
We see that our clustering is outperforming D\&M by a comfortable margin: $+3.7$ PQ pts. With box-splitting, \ours would gain $+0.4$ further PQ pts.

\begin{table}[t]
    \centering
    \small
    \setlength{\tabcolsep}{3pt}
    
    \begin{tabular}{l|rrrr}
    \toprule
    Method & \multicolumn{1}{c}{PQ\!\!} & \multicolumn{1}{c}{RQ\!\!} & \multicolumn{1}{c}{SQ\!\!} & \multicolumn{1}{c}{mIoU\!\!} \\
    \midrule
    3DUIS & 60.2 & 70.7 & 79.7 & 72.2 \\
    UNIT  & 61.1 & 72.0 & 79.0 & 72.2 \\
    \ours & 65.9 & 75.5 & 81.4 & 72.2 \\
    \bottomrule
    \end{tabular}
    \caption{\textbf{Comparison to semantic-free instance segmentations} evaluated on the SemanticKITTI validation set.}
    \label{tab:comparisons_unsupervised}
\end{table}

\paragraph{Comparison to semantic-free instance segmentation.}
3DUIS~\cite{3duis} and UNIT~\cite{unit} are two unsupervised instance segmentation methods. They propose to find individual instances of objects in a totally unsupervised manner, regardless of semantics. While their task is more involved, as they cannot rely on predicted semantics mask, we can still compare their instance predictions with that of \ours. In practice, we use the intersection of their instance mask and each semantic mask as panoptic predictions. As can be noted in \cref{tab:comparisons_unsupervised}, \ours significantly outperforms those methods. It can still be noted that both unsupervised method work quite impressively, almost reaching the level of performance of D\&M while not using any semantic information.

    

\begin{table}[t]
    \centering
    \small
    \setlength{\tabcolsep}{3pt}
    
    \begin{tabular}{ll|rrrr}
    \toprule
    Sem.\,Seg & Inst.\,Seg & \multicolumn{1}{c}{PQ\!\!} & \multicolumn{1}{c}{RQ\!\!} & \multicolumn{1}{c}{SQ\!\!} & \multicolumn{1}{c}{mIoU\!\!} \\
    \midrule
    \emph{Oracle} & \ours & 96.2 & 98.4 & 97.8 & 100.0 \\
    WI-768 & \emph{Oracle} & 81.7 & 88.1 & 94.1 & 81.4 \\
    PTv3 & \emph{Oracle} & 83.2 & 87.6 & 92.8 & 81.5 \\
    \emph{Oracle} & \emph{Oracle} & 100.0 & 100.0 & 100.0 & 100.0 \\
    \bottomrule
    \end{tabular}
    \caption{\textbf{Oracle results} evaluated on the nuScenes validation set.}
    \label{tab:oracles}
\end{table}

\subsection{Evaluation of the clustering upper bound}
\label{sec:oracles}

To evaluate what remains to be gained by developing better semantics or instance heads, we devise an oracle head for both subtasks: a ``semantic oracle'' using the ground-truth semantic segmentation as semantic predictions, and an ``instance oracle'' splitting the semantic predictions of a given method using the
ground-truth instance boundaries, without affecting the semantic prediction.
Results, in \cref{tab:oracles}, shows how much scores are expected to improve with both oracles. 
When using the ``semantic oracle'', the PQ on nuScenes raises to 96.2, (and 99.0 on SemanticKITTI), while using PTv3 predictions and the instance oracle only raises the PQ by +4.3 to 83.2, showing that instance extraction is much more saturated than semantic segmentation.

\begin{table}[t]
    \centering
    \small
    \setlength{\tabcolsep}{4.1pt}
    
    \begin{tabular}{ccc |ccc}
    \toprule
    \multirow{2}{*}{BEV} & Box & \multirow{2}{*}{Threshold} & \multirow{2}{*}{PQ} & \multirow{2}{*}{PQ\textsuperscript{$\dagger$}} & \multirow{2}{*}{mIoU} \\
    & splitting &  & & & \\
    \midrule
    \xmark & \xmark & const.           & 76.3 & 78.7 & 81.5\\
    \cmark & \xmark & const.           & 76.9 & 79.3 & 81.5\\
    \xmark & \cmark & const.           & 77.1 & 79.5 & 81.5\\
    \rowcolor{green!15}
    \cmark & \cmark & const.           & 78.9 & 81.3 & 81.5\\
    \midrule
    \xmark & \xmark & $\propto |d|$ \cite{less}    & 75.9 & 78.3 & 81.5\\

    \bottomrule
    \end{tabular}
    \caption{\textbf{Ablation study} of our different clustering components on the nuScenes validation set. $|d|$ is the distance to the sensor.}
    \label{tab:ablation}
\end{table}

\begin{table}[t]
    \small
    \centering
    \setlength{\tabcolsep}{2.5pt}    
    \begin{tabular}{l@{~}|c|ccc|ccc}
    
    \multirow{2}{*}{\centering Method}
    & sem. & \multicolumn{3}{c|}{statistics-based} &  \multicolumn{3}{c}{web-based}\\
    
    & mIoU & PQ & PQ\textsuperscript{$\dagger$} & PQ\textsubscript{Th} & PQ & PQ\textsuperscript{$\dagger$} & PQ\textsubscript{Th}
    \\
    
    \midrule
    \multicolumn{8}{l}{\textit{SemanticKITTI}}\\
    MinkUNet w/o TTA & 70.7 & 64.1 & 68.8 & 72.5 & 64.2 & 68.9 & 72.6 \\
    PTv3 & 67.3 & 62.4 & 66.1 & 65.9 & 62.4 & 66.2 & 66.1 \\
    WI-256 & 70.3 & 64.0 & 68.9 & 69.0 & 64.2 & 69.0 & 69.3\\
    MinkUNet & 72.2 & 65.8 & 70.1 & 73.7 & 65.8 & 70.1 & 73.7 \\
    MinkUNet \& PTv3 & 72.0 & 66.5 & 70.7 & 73.2 & 66.6 & 70.8 & 73.4 \\

    \midrule
    \multicolumn{8}{l}{\textit{nuScenes}}\\
    WI-768 w/o TTA & 80.3 & 77.2 & 80.1 & 78.3 & 76.9 & 89.9 & 77.9 \\
    WI-768 & 81.4 & 78.2 & 81.0 & 79.0 & 77.9 & 80.7 & 78.6 \\
    PTv3 & 81.5 & 79.0 & 81.4 & 79.6 & 78.9 & 81.3 & 79.3 \\
    WI-768 \& PTv3 & 82.7 & 79.8 & 82.2 & 80.5 & 79.5 & 81.9 & 80.2 \\
    \midrule
    \multicolumn{8}{l}{\textit{SemanticPOSS}}\\
    PTv3 & 58.3 & 51.1 & 57.4 & 63.6 & 51.4 & 57.7 & 64.8 \\

    \end{tabular}

    \vspace{-0.2cm}
    \caption{\textbf{Comparison between statistics-based and web-based parameters settings.} 
    Run details can be found in \cref{tab:results_nuscenes_val_compressed,tab:results_semkitti_val_compressed}.
    }
    \label{tab:ablation_web}
\end{table}

\subsection{Ablation and sensitivity study}
\label{sec:ablation}

\paragraph{Influence of each \ours modules.}
We present our ablation study in \cref{tab:ablation}, conducted on the validation set of nuScenes, using PTv3 as semantic backbone. We see that constructing the graph in BEV or post-processing the result with our box splitting strategy improve the results: respectively $+0.6$ point and $+0.8$ point in PQ over the baseline.
The combination of both improve the PQ by $2.6$ points.

Instead of using a constant threshold to keep or remove the edge between two points when constructing our graph, we also tested a threshold varying linearly with the range, as used in LESS~\cite{less}. The linear coefficient is the ALPINE threshold $t_c$ multiplied by a constant optimized at dataset scale to maximize performance. Despite this optimization, we find that, in our case, using a constant threshold depending on the typical size of the objects leads to better results.

\paragraph{Dataset-based vs. web-based results.}

\cref{tab:ablation_web} gives the difference in scores for the different sets of parameters described in \cref{sec:parameters}. Both choices of parameters give very similar results, from which we conclude \ours parameters are both easy to setup, have a sensible meaning, and do not require annotations.



\section{Limitations and scope of this article}

Since our method has a single fixed threshold per class, failure cases can be crafted by making two objects closer than the chosen threshold.
It would thus be tempting to build intricate training-based solutions, for example  by learning the neighboring condition~\cite{gps3net,superpointgraph}.

However, \ours proves that old techniques 
can reach state-of-the-art results, with no need for any object label, tuning, learning or extremely complex designs. 
\ours is a very strong baseline that future methods will have to beat to prove that their design actually benefits the task.

Furthermore, this object resolution limitation naturally reduces as the point density (or equivalently the LiDAR angular resolution) increases and intra-object average points distances reduces, leading to believe this limitation could mostly solve itself with new LiDAR technologies.

\section{Conclusion}

\ours is reaching state-of-the-art results 
in LiDAR panoptic segmentation with no instance labels, no training, and no heavy computation requiring a GPU. Furthermore, \ours can be taken off-the-shelf and applied as is on top of any semantic predictor.
Our study of \ours in replacement of prior instance predictions' heads proves that instance extraction is largely saturated.



{This work shows}
that (1) 
{instance} labels are not {fully exploited}
by current LiDAR panoptic segmentation methods and (2) simple clustering {as ALPINE} should always be tested as a baseline in order to assess the gain {obtained with trainable} panoptic heads.
We also release our code for integration into any semantic segmentation backbone for panoptic segmentation.

{
    \small
    \bibliographystyle{ieeenat_fullname}
    \bibliography{main}

@String(PAMI = {IEEE Trans. Pattern Anal. Mach. Intell.})

@String(CVPR= {IEEE Conf. Comput. Vis. Pattern Recog.})

@String(ICCV= {Int. Conf. Comput. Vis.})

@String(ECCV= {Eur. Conf. Comput. Vis.})

@String(NIPS= {Adv. Neural Inform. Process. Syst.})

@String(BMVC= {Brit. Mach. Vis. Conf.})

@String(ICASSP=	{ICASSP})

@String(AAAI = {AAAI})

@String(PAMI  = {IEEE TPAMI})

@String(CVPR  = {CVPR})

@String(ICCV  = {ICCV})

@String(ECCV  = {ECCV})

@String(NIPS  = {NeurIPS})

@String(BMVC  =	{BMVC})

@inproceedings{uniseg,
    author={Liu, Youquan and Chen, Runnan and Li, Xin and Kong, Lingdong and Yang, Yuchen and Xia, Zhaoyang and Bai, Yeqi and Zhu, Xinge and Ma, Yuexin and Li, Yikang and Qiao, Yu and Hou, Yuenan},
    booktitle={ICCV}, 
    title={{UniSeg}: A Unified Multi-Modal LiDAR Segmentation Network and the OpenPCSeg Codebase}, 
    year={2023},
}

@inproceedings{phnet,
  title={{Panoptic-PHNet}: Towards real-time and high-precision {LiDAR} panoptic segmentation via clustering pseudo heatmap},
  author={Li, Jinke and He, Xiao and Wen, Yang and Gao, Yuan and Cheng, Xiaoqiang and Zhang, Dan},
  booktitle={CVPR},
  year={2022}
}

@article{maskpls,
  title={Mask-based panoptic {LiDAR} segmentation for autonomous driving},
  author={Marcuzzi, Rodrigo and Nunes, Lucas and Wiesmann, Louis and Behley, Jens and Stachniss, Cyrill},
  journal={IEEE Robotics and Automation Letters},
  volume={8},
  number={2},
  pages={1141--1148},
  year={2023},
}

@inproceedings{panopticpolarnet,
  title={Panoptic-polarnet: Proposal-free {LiDAR} point cloud panoptic segmentation},
  author={Zhou, Zixiang and Zhang, Yang and Foroosh, Hassan},
  booktitle={CVPR},
  year={2021}
}

@inproceedings{dsnet,
  title={{LiDAR}-based panoptic segmentation via dynamic shifting network},
  author={Hong, Fangzhou and Zhou, Hui and Zhu, Xinge and Li, Hongsheng and Liu, Ziwei},
  booktitle={CVPR},
  year={2021}
}

@inproceedings{cpseg,
  title={{CPSeg}: Cluster-free panoptic segmentation of 3d {LiDAR} point clouds},
  author={Li, Enxu and Razani, Ryan and Xu, Yixuan and Liu, Bingbing},
  booktitle={ICRA},
  year={2023},
}

@article{efficientlps,
  title={{EfficientLPS}: Efficient {LiDAR} panoptic segmentation},
  author={Sirohi, Kshitij and Mohan, Rohit and B{\"u}scher, Daniel and Burgard, Wolfram and Valada, Abhinav},
  journal={IEEE Transactions on Robotics},
  volume={38},
  number={3},
  pages={1894--1914},
  year={2021},
}

@inproceedings{gps3net,
  title={{GP-S3Net}: Graph-based panoptic sparse semantic segmentation network},
  author={Razani, Ryan and Cheng, Ran and Li, Enxu and Taghavi, Ehsan and Ren, Yuan and Bingbing, Liu},
  booktitle={ICCV},
  year={2021}
}

@inproceedings{pups,
  title={{PUPS}: Point cloud unified panoptic segmentation},
  author={Su, Shihao and Xu, Jianyun and Wang, Huanyu and Miao, Zhenwei and Zhan, Xin and Hao, Dayang and Li, Xi},
  booktitle={AAAI},
  year={2023}
}

@inproceedings{lidarmultinet,
  title={Lidarmultinet: Towards a unified multi-task network for {LiDAR} perception},
  author={Ye, Dongqiangzi and Zhou, Zixiang and Chen, Weijia and Xie, Yufei and Wang, Yu and Wang, Panqu and Foroosh, Hassan},
  booktitle={AAAI},
  year={2023}
}

@inproceedings{tornadonet,
  title={Tornado-net: multiview total variation semantic segmentation with diamond inception module},
  author={Gerdzhev, Martin and Razani, Ryan and Taghavi, Ehsan and Bingbing, Liu},
  booktitle={ICRA},
  year={2021},
}

@inproceedings{spvcnnpp,
  title={Searching efficient 3d architectures with sparse point-voxel convolution},
  author={Tang, Haotian and Liu, Zhijian and Zhao, Shengyu and Lin, Yujun and Lin, Ji and Wang, Hanrui and Han, Song},
  booktitle={ECCV},
  year={2020},
}

@article{mopt,
  title={Mopt: Multi-object panoptic tracking},
  author={Hurtado, Juana Valeria and Mohan, Rohit and Burgard, Wolfram and Valada, Abhinav},
  journal={arXiv preprint arXiv:2004.08189},
  year={2020}
}

@inproceedings{lidarpanoptic,
  title={{LiDAR} panoptic segmentation for autonomous driving},
  author={Milioto, Andres and Behley, Jens and McCool, Chris and Stachniss, Cyrill},
  booktitle={IROS},
  year={2020},
}

@inproceedings{smacseg,
  title={{SMAC-Seg}: {LiDAR} panoptic segmentation via sparse multi-directional attention clustering},
  author={Li, Enxu and Razani, Ryan and Xu, Yixuan and Liu, Bingbing},
  booktitle={ICRA},
  year={2022},
}

@article{panoster,
  title={Panoster: End-to-end panoptic segmentation of {LiDAR} point clouds},
  author={Gasperini, Stefano and Mahani, Mohammad-Ali Nikouei and Marcos-Ramiro, Alvaro and Navab, Nassir and Tombari, Federico},
  journal={IEEE Robotics and Automation Letters},
  volume={6},
  number={2},
  pages={3216--3223},
  year={2021},
}

@inproceedings{kpconv,
  title={{KPConv}: Flexible and deformable convolution for point clouds},
  author={Thomas, Hugues and Qi, Charles R and Deschaud, Jean-Emmanuel and Marcotegui, Beatriz and Goulette, Fran{\c{c}}ois and Guibas, Leonidas J},
  booktitle={CVPR},
  year={2019}
}

@inproceedings{pointpillars,
  title={Pointpillars: Fast encoders for object detection from point clouds},
  author={Lang, Alex H and Vora, Sourabh and Caesar, Holger and Zhou, Lubing and Yang, Jiong and Beijbom, Oscar},
  booktitle={CVPR},
  year={2019}
}

@inproceedings{cylinder3d,
  title={Cylindrical and asymmetrical 3d convolution networks for {LiDAR} segmentation},
  author={Zhu, Xinge and Zhou, Hui and Wang, Tai and Hong, Fangzhou and Ma, Yuexin and Li, Wei and Li, Hongsheng and Lin, Dahua},
  booktitle={CVPR},
  year={2021}
}

@article{polarstream,
  title={Polarstream: Streaming object detection and segmentation with polar pillars},
  author={Chen, Qi and Vora, Sourabh and Beijbom, Oscar},
  journal={NeurIPS},
  volume={34},
  year={2021}
}

@inproceedings{scan,
  title={Sparse cross-scale attention network for efficient {LiDAR} panoptic segmentation},
  author={Xu, Shuangjie and Wan, Rui and Ye, Maosheng and Zou, Xiaoyi and Cao, Tongyi},
  booktitle={AAAI},
  volume={36},
  number={3},
  pages={2920--2928},
  year={2022}
}

@inproceedings{pvcl,
  title={Prototype-voxel contrastive learning for {LiDAR} point cloud panoptic segmentation},
  author={Liu, Minzhe and Zhou, Qiang and Zhao, Hengshuang and Li, Jianing and Du, Yuan and Keutzer, Kurt and Du, Li and Zhang, Shanghang},
  booktitle={ICRA},
  year={2022},
}

@article{dsnet2,
  title={Unified 3d and 4d panoptic segmentation via dynamic shifting networks},
  author={Hong, Fangzhou and Kong, Lingdong and Zhou, Hui and Zhu, Xinge and Li, Hongsheng and Liu, Ziwei},
  journal={PAMI},
  year={2024},
}

@InProceedings{lcps,
    author    = {Zhang, Zhiwei and Zhang, Zhizhong and Yu, Qian and Yi, Ran and Xie, Yuan and Ma, Lizhuang},
    title     = {{LiDAR}-Camera Panoptic Segmentation via Geometry-Consistent and Semantic-Aware Alignment},
    booktitle = {ICCV},
    year      = {2023},
}

@inproceedings{vin,
  title={{VIN}: Voxel-based implicit network for joint 3d object detection and segmentation for lidars},
  author={Zhong, Yuanxin and Zhu, Minghan and Peng, Huei},
  booktitle={BMVC},
  year={2021}
}

@inproceedings{pointgroup,
  title={Pointgroup: Dual-set point grouping for 3d instance segmentation},
  author={Jiang, Li and Zhao, Hengshuang and Shi, Shaoshuai and Liu, Shu and Fu, Chi-Wing and Jia, Jiaya},
  booktitle={CVPR},
  year={2020}
}

@article{locationguided,
  title={Location-guided {LiDAR}-based panoptic segmentation for autonomous driving},
  author={Xian, Guozeng and Ji, Changyun and Zhou, Lin and Chen, Guang and Zhang, Junping and Li, Bin and Xue, Xiangyang and Pu, Jian},
  journal={IEEE Transactions on Intelligent Vehicles},
  volume={8},
  number={2},
  pages={1473--1483},
  year={2022},
}

@inproceedings{pointrcnn,
  title={Pointrcnn: 3d object proposal generation and detection from point cloud},
  author={Shi, Shaoshuai and Wang, Xiaogang and Li, Hongsheng},
  booktitle={CVPR},
  year={2019}
}

@inproceedings{polarnet,
  title={Polarnet: An improved grid representation for online {LiDAR} point clouds semantic segmentation},
  author={Zhang, Yang and Zhou, Zixiang and David, Philip and Yue, Xiangyu and Xi, Zerong and Gong, Boqing and Foroosh, Hassan},
  booktitle={CVPR},
  year={2020}
}

@inproceedings{less,
  title={{LESS}: Label-efficient semantic segmentation for {LiDAR} point clouds},
  author={Liu, Minghua and Zhou, Yin and Qi, Charles R and Gong, Boqing and Su, Hao and Anguelov, Dragomir},
  booktitle={ECCV},
  year={2022},
}

@inproceedings{panet,
  title={{PANet}: {LiDAR} Panoptic Segmentation with Sparse Instance Proposal and Aggregation},
  author={Mei, Jianbiao and Yang, Yu and Wang, Mengmeng and Hou, Xiaojun and Li, Laijian and Liu, Yong},
  booktitle={IROS},
  year={2023},
}

@misc{openpcdet,
    title={OpenPCDet: An Open-source Toolbox for 3D Object Detection from Point Clouds},
    author={OpenPCDet Development Team},
    howpublished = {\url{https://github.com/open-mmlab/OpenPCDet}},
    year={2020}
}

@inproceedings{cfnet,
  title={Center Focusing Network for Real-Time {LiDAR} Panoptic Segmentation},
  author={Li, Xiaoyan and Zhang, Gang and Wang, Boyue and Hu, Yongli and Yin, Baocai},
  booktitle={CVPR},
  year={2023}
}

@inproceedings{centerlps,
  title={{CenterLPS}: Segment instances by centers for {LiDAR} panoptic segmentation},
  author={Mei, Jianbiao and Yang, Yu and Wang, Mengmeng and Li, Zizhang and Hou, Xiaojun and Ra, Jongwon and Li, Laijian and Liu, Yong},
  booktitle={Proceedings of the 31st ACM International Conference on Multimedia},
  pages={1884--1894},
  year={2023}
}

@article{ieqlps,
  title={Instance Embedding as Queries for DETR-based {LiDAR} Panoptic Segmentation},
  author={Ha-Phan, Ngoc-Quan and Vuong, Hung Viet and Yoo, Myungsik},
  journal={IEEE Transactions on Intelligent Vehicles},
  year={2024},
}

@INPROCEEDINGS{dqformer,
  author={Yang, Hao and Lin, Shuyuan and Jiang, Runqing and Lu, Yang and Wang, Hanzi},
  booktitle={ICASSP 2023 - 2023 IEEE International Conference on Acoustics, Speech and Signal Processing (ICASSP)}, 
  title={DQFORMER: Dynamic Query Transformer for Lane Detection}, 
  year={2023},
  pages={1-5},
  doi={10.1109/ICASSP49357.2023.10097047}}

@inproceedings{aopnet,
  title={{AOP-Net}: All-in-One Perception Network for {LiDAR}-based Joint 3D Object Detection and Panoptic Segmentation},
  author={Xu, Yixuan and Fazlali, Hamidreza and Ren, Yuan and Liu, Bingbing},
  booktitle={2023 IEEE Intelligent Vehicles Symposium (IV)},
  year={2023},
}

@article{p3former,
  title={Position-guided point cloud panoptic segmentation transformer},
  author={Xiao, Zeqi and Zhang, Wenwei and Wang, Tai and Loy, Chen Change and Lin, Dahua and Pang, Jiangmiao},
  journal={International Journal of Computer Vision},
  pages={1--16},
  year={2024},
}

@InProceedings{pointnext,
    author = {Qian, Guocheng and Li, Yuchen and Peng, Houwen and Mai, Jinjie and Hammoud, Hasan Abed Al Kader and Elhoseiny, Mohamed and Ghanem, Bernard},
    title = {{PointNeXt: Revisiting PointNet++ with Improved Training and Scaling Strategies}},
    booktitle = NIPS,
    year = {2022},
}

@InProceedings{pointtransformer,
    author    = {Zhao, Hengshuang and Jiang, Li and Jia, Jiaya and Torr, Philip H.S. and Koltun, Vladlen},
    title     = {{Point Transformer}},
    booktitle = ICCV,
    year      = {2021},
}

@InProceedings{pointmixer,
    author={Choe, Jaesung and Park, Chunghyun and Rameau, Francois and Park, Jaesik and Kweon, In So},
    title={{PointMixer: MLP-Mixer for Point Cloud Understanding}},
    booktitle=ECCV,
    year="2022",
}

@InProceedings{pointnet,
    author = {Qi, Charles R. and Su, Hao and Mo, Kaichun and Guibas, Leonidas J.},
    title = {{PointNet: Deep Learning on Point Sets for 3D Classification and Segmentation}},
    booktitle = CVPR,
    year = {2017}
}

@inproceedings{pointnetpp,
    author = {Qi, Charles Ruizhongtai and Yi, Li and Su, Hao and Guibas, Leonidas J},
    booktitle = NIPS,
    title = {{PointNet++: Deep Hierarchical Feature Learning on Point Sets in a Metric Space}},
    year = {2017}
}

@InProceedings{rangeformer,
      title={Rethinking Range View Representation for LiDAR Segmentation}, 
      author={Lingdong Kong and Youquan Liu and Runnan Chen and Yuexin Ma and Xinge Zhu and Yikang Li and Yuenan Hou and Yu Qiao and Ziwei Liu},
      year={2023},
      booktitle = ICCV,
}

@InProceedings{sdseg3d,
    author={Jiale Li and Hang Dai and Yong Ding},
    title={Self-Distillation for Robust LiDAR Semantic Segmentation in Autonomous Driving},
    booktitle = {ECCV},
    year=2022,
}

@InProceedings{RPVNet,
    author    = {Xu, Jianyun and Zhang, Ruixiang and Dou, Jian and Zhu, Yushi and Sun, Jie and Pu, Shiliang},
    title     = {{RPVNet: A Deep and Efficient Range-Point-Voxel Fusion Network for LiDAR Point Cloud Segmentation}},
    booktitle = ICCV,
    year      = {2021},
}

@article{gfnet,
    author={Haibo Qiu and Baosheng Yu and Dacheng Tao},
    title={{GFNet: Geometric Flow Network for 3D Point Cloud Semantic Segmentation}},    
    journal={Transactions on Machine Learning Research},
    year={2022},
    url={https://openreview.net/forum?id=LSAAlS7Yts},
    note={}
}

@inproceedings{bevcontrast,
  author    = {Corentin Sautier and Gilles Puy and Alexandre Boulch and Renaud Marlet and Vincent Lepetit},
  title     = {{BEVContrast}: Self-Supervision in BEV Space for Automotive Lidar Point Clouds},
  booktitle = {3DV},
  year      = 2024
}

@InProceedings{af2s3net,
    author    = {Cheng, Ran and Razani, Ryan and Taghavi, Ehsan and Li, Enxu and Liu, Bingbing},
    title     = {{(AF)2-S3Net: Attentive Feature Fusion With Adaptive Feature Selection for Sparse Semantic Segmentation Network}},
    booktitle = CVPR,
    year      = {2021},
}

@InProceedings{SqueezeSegV3,
    author="Xu, Chenfeng
    and Wu, Bichen
    and Wang, Zining
    and Zhan, Wei
    and Vajda, Peter
    and Keutzer, Kurt
    and Tomizuka, Masayoshi",
    editor="Vedaldi, Andrea
    and Bischof, Horst
    and Brox, Thomas
    and Frahm, Jan-Michael",
    title={{SqueezeSegV3: Spatially-Adaptive Convolution for Efficient Point-Cloud Segmentation}},
    booktitle=ECCV,
    year="2020",
}

@INPROCEEDINGS{SalsaNet,
    author={Aksoy, Eren Erdal and Baci, Saimir and Cavdar, Selcuk},
    booktitle={IEEE Intelligent Vehicles Symposium (IV)}, 
    title={{SalsaNet: Fast Road and Vehicle Segmentation in LiDAR Point Clouds for Autonomous Driving}}, 
    year={2020},
}

@INPROCEEDINGS{rangenetpp,
    author={Milioto, Andres and Vizzo, Ignacio and Behley, Jens and Stachniss, Cyrill},
    booktitle={IROS}, 
    title={{RangeNet ++: Fast and Accurate LiDAR Semantic Segmentation}}, 
    year={2019},
}

@inproceedings{ptv2,
  title={Point Transformer V2: Grouped Vector Attention and Partition-based Pooling},
  author={Wu, Xiaoyang and Lao, Yixing and Jiang, Li and Liu, Xihui and Zhao, Hengshuang},
  booktitle={NeurIPS},
  year={2022}
}

@inproceedings{minkunet,
  author = {Choy, Christopher and Gwak, JunYoung and Savarese, Silvio},
  title = {{{4D} Spatio-Temporal {ConvNets}: Minkowski Convolutional Neural Networks}},
  booktitle = CVPR,
  year = 2019
}

@inproceedings{waffleiron,
  title={Using a Waffle Iron for Automotive Point Cloud Semantic Segmentation},
  author={Puy, Gilles and Boulch, Alexandre and Marlet, Renaud},
  booktitle={ICCV},
  year={2023}
}

@misc{openpcseg,
    title={{OpenPCSeg}: An Open Source Point Cloud Segmentation Codebase},
    author={Liu, Youquan and Bai, Yeqi and Kong, Lingdong and Chen, Runnan and Hou, Yuenan and Shi, Botian and Li, Yikang},
    howpublished = {\url{https://github.com/PJLab-ADG/PCSeg}},
    year={2023}
}

@inproceedings{ptv3,
    title={Point Transformer V3: Simpler, Faster, Stronger},
    author={Wu, Xiaoyang and Jiang, Li and Wang, Peng-Shuai and Liu, Zhijian and Liu, Xihui and Qiao, Yu and Ouyang, Wanli and He, Tong and Zhao, Hengshuang},
    booktitle={CVPR},
    year={2024}
}

@article{kdtree,
  title={Multidimensional binary search trees used for associative searching},
  author={Bentley, Jon Louis},
  journal={Communications of the ACM},
  volume={18},
  number={9},
  pages={509--517},
  year={1975},
  publisher={ACM New York, NY, USA}
}

@inproceedings{nuscenes,
  author = {Caesar, Holger and Bankiti, Varun and Lang, Alex H. and Vora, Sourabh and Liong, Venice Erin and Xu, Qiang and Krishnan, Anush and Pan, Yu and Baldan, Giancarlo and Beijbom, Oscar},
  title = {{{nuScenes}: A Multimodal Dataset for Autonomous Driving}},
  booktitle = CVPR,
  year = 2020
}

@inproceedings{semantickitti,
  author = {Behley, Jens and Garbade, Martin and Milioto, Andres and Quenzel, Jan and Behnke, Sven and Stachniss, Cyrill and Gall, Jurgen},
  title = {{{SemanticKITTI}: A Dataset for Semantic Scene Understanding of {LiDAR} Sequences}},
  booktitle = ICCV,
  year = 2019
}

@inproceedings{semanticposs,
  title={{SemanticPOSS}: A point cloud dataset with large quantity of dynamic instances},
  author={Pan, Yancheng and Gao, Biao and Mei, Jilin and Geng, Sibo and Li, Chengkun and Zhao, Huijing},
  booktitle={2020 IEEE Intelligent Vehicles Symposium (IV)},
  year={2020},
}

@article{second,
  title={{SECOND}: Sparsely Embedded Convolutional Detection},
  author={Yan Yan and Yuxing Mao and Bo Li},
  journal={Sensors},
  volume={18},
  year={2018}
}

@inproceedings{lpst,
  title={{LiDAR} Panoptic Segmentation and Tracking without Bells and Whistles},
  author={Agarwalla, Abhinav and Huang, Xuhua and Ziglar, Jason and Ferroni, Francesco and Leal-Taix{\'e}, Laura and Hays, James and O{\v{s}}ep, Aljo{\v{s}}a and Ramanan, Deva},
  booktitle={IROS},
  year={2023},
}

@inproceedings{dividemerge,
  title={A divide-and-merge point cloud clustering algorithm for {LiDAR} panoptic segmentation},
  author={Zhao, Yiming and Zhang, Xiao and Huang, Xinming},
  booktitle={ICRA},
  year={2022},
}

@inproceedings{technicalsurvey,
  title={A technical survey and evaluation of traditional point cloud clustering methods for {LiDAR} panoptic segmentation},
  author={Zhao, Yiming and Zhang, Xiao and Huang, Xinming},
  booktitle=ICCV,
  year={2021}
}

@inproceedings{slr,
  title={Fast segmentation of 3d point clouds: A paradigm on {LiDAR} data for autonomous vehicle applications},
  author={Zermas, Dimitris and Izzat, Izzat and Papanikolopoulos, Nikolaos},
  booktitle={ICRA},
  year={2017},
}

@article{maskrange,
  title={Maskrange: A mask-classification model for range-view based {LiDAR} segmentation},
  author={Gu, Yi and Huang, Yuming and Xu, Chengzhong and Kong, Hui},
  journal={arXiv preprint arXiv:2206.12073},
  year={2022}
}

@inproceedings{maskformer,
  title={Per-Pixel Classification is Not All You Need for Semantic Segmentation},
  author={Bowen Cheng and Alexander G. Schwing and Alexander Kirillov},
  booktitle={NeurIPS},
  year={2021}
}

@inproceedings{rangeviT,
  title={{RangeViT}: Towards Vision Transformers for 3D Semantic Segmentation in Autonomous Driving},
  author={Ando, Angelika and Gidaris, Spyros and Bursuc, Andrei and Puy, Gilles and Boulch, Alexandre and Marlet, Renaud},
  booktitle={CVPR},
  year={2023}
}

@article{3duis,
  author = {Nunes, Lucas and Chen, Xieyuanli and Marcuzzi, Rodrigo and Osep, Aljosa and Leal-Taix\'e, Laura and Stachniss, Cyrill and Behley, Jens},
  title = {{Unsupervised {Class}-{Agnostic} Instance Segmentation of {3D} {LiDAR} Data for Autonomous Vehicles}},
  journal = {IEEE Robotics and Automation Letters (RA-L)},
  year = 2022
}

@inproceedings{tarl,
  author = {Nunes, Lucas and Wiesmann, Louis and Marcuzzi, Rodrigo and Chen, Xieyuanli and Behley, Jens and Stachniss, Cyrill},
  title = {{Temporal Consistent {3D} {LiDAR} Representation Learning for Semantic Perception in Autonomous Driving}},
  booktitle = CVPR,
  year = 2023
}

@inproceedings{unit,
  title = {{UNIT}: Unsupervised Online Instance Segmentation through Time},
  author = {Corentin Sautier and Gilles Puy and Alexandre Boulch and Renaud Marlet and Vincent Lepetit},
  booktitle={3DV},
  year = {2025}
}

@article{rusu2010semantic,
  title={Semantic 3D object maps for everyday manipulation in human living environments},
  author={Rusu, Radu Bogdan},
  journal={KI-K{\"u}nstliche Intelligenz},
  volume={24},
  pages={345--348},
  year={2010},
}

@InProceedings{pcl,
  author    = {Radu Bogdan Rusu and Steve Cousins},
  title     = {{3D is here: Point Cloud Library (PCL)}},
  booktitle = {ICRA},
  year      = {2011},
}

@article{slic,
  title={{SLIC} superpixels compared to state-of-the-art superpixel methods},
  author={Achanta, Radhakrishna and Shaji, Appu and Smith, Kevin and Lucchi, Aurelien and Fua, Pascal and S{\"u}sstrunk, Sabine},
  journal={TPAMI},
  year={2012},
}

@inproceedings{modest,
  title = {Learning to Detect Mobile Objects from LiDAR Scans Without Labels},
  author = {You, Yurong and Luo, Katie Z and Phoo, Cheng Perng and Chao, Wei-Lun and Sun, Wen and Hariharan, Bharath and Campbell, Mark and Weinberger, Kilian Q.},
  booktitle = CVPR,
  year = {2022},
  month = jun
}

@inproceedings{superpointgraph,
  title={Scalable 3D panoptic segmentation as superpoint graph clustering},
  author={Robert, Damien and Raguet, Hugo and Landrieu, Loic},
  booktitle={3DV},
  year={2024},
}

@misc{stackoverflow,
    title = {Algorithm to find the minimum-area-rectangle for given points in order to compute the major and minor axis length},
    author = {Gianni Spear},
    howpublished = {\url{https://stackoverflow.com/questions/13542855/algorithm-to-find-the-minimum-area-rectangle-for-given-points-in-order-to-comput}},
    year = {2012},
}

@misc{carsize_US,
    title = {Supersized: A Decade-Long Growth of {U.S.} Cars Reveals a Bigger Picture},
    author = {FINN},
    howpublished = {\url{https://www.finn.com/en-DE/campaign/supersized}},
    year = {2024}
}

@misc{carsize_EUR,
    title = {Average Vehicle Size In The {US} And Europe Is Larger Than Ever},
    author = {Motor1},
    howpublished = {\url{https://www.motor1.com/news/707996/vehicles-larger-than-ever-usa-europe/}},
    year = {2024}
}

@misc{bikesize,
    title = {One Bike Average Size},
    author = {thebestbikelock.com},
    howpublished = {\url{https://thebestbikelock.com/wp-content/uploads/2020/01/one-bike-average-size.gif}},
    year = {2020}
}

@misc{motorcyclesize,
    title = {What is the average size of a motorbike},
    author = {carparkjourney.wordpress.com},
    howpublished = {\url{https://carparkjourney.wordpress.com/2013/07/16/what-is-the-average-size-of-a-motorbike/}},
    year = {2013}
}

@misc{height,
    title = {Average human height by country},
    author = {Wikipedia},
    howpublished = {\url{https://en.wikipedia.org/wiki/Average_human_height_by_country}},
    year = {2024}
}

@inproceedings{armspan,
    title = {All-age relationship between arm span and height in different ethnic groups},
    author = {Quanjer, Philip H and Capderou, André and Mazicioglu, Mumtaz M and Aggarwal Ashutosh N and Banik Sudip Datta and Popovic Stevo and Tayie Francis A and Golshan Mohammad, Ip Mary S M, Zelter Marc},
    booktitle = {European Respiratory Journal},
    year = {2014}
}

@INPROCEEDINGS{wang2020trainingermany,
  author={Wang, Yan and Chen, Xiangyu and You, Yurong and Li, Li Erran and Hariharan, Bharath and Campbell, Mark and Weinberger, Kilian Q. and Chao, Wei-Lun},
  booktitle=CVPR, 
  title={Train in {Germany}, Test in the {USA}: Making {3D} Object Detectors Generalize}, 
  year={2020},
}

@inproceedings{rbnn,
  title={A clustering method for efficient segmentation of 3D laser data},
  author={Klasing, Klaas and Wollherr, Dirk and Buss, Martin},
  booktitle={ICRA},
  year={2008},
}

@inproceedings{supervoxels,
  title={Voxel cloud connectivity segmentation-supervoxels for point clouds},
  author={Papon, Jeremie and Abramov, Alexey and Schoeler, Markus and Worgotter, Florentin},
  booktitle={CVPR},
  year={2013}
}

@inproceedings{clustering_1,
  title={Fast range image-based segmentation of sparse 3D laser scans for online operation},
  author={Bogoslavskyi, Igor and Stachniss, Cyrill},
  booktitle={IROS},
  year={2016},
}
}

\clearpage
\maketitlesupplementary
\appendix

\section{Margin of improvement with oracle analysis}





\subsection{Discussion}

Results in \cref{sec:oracles} motivate our conclusion that instance labels are mostly unnecessary or not correctly utilized for outdoor LiDAR panoptic segmentation.

On the other hand, we observe that the only methods competing with \ours on SemanticKITTI and nuScenes are end-to-end, query-based methods.
This might hint that there is more to benefit from and end-to-end panoptic training than those two-staged oracles would suggest, for which we show in \cref{tab:oracles} that there is little gain left to expect with better instance predictions.
Indeed end-to-end methods, and in particular query-based ones, benefit from the simultaneous instance and semantic segmentation tasks training.
However, based on the tables of the main paper, the gain remains currently small, if positive, compared to our instance-annotation-free approach.
This could hint at an existing margin for improvement for end-to-end methods.

\section{Metrics}

In LiDAR point cloud segmentation, two primary metrics are used for evaluation: mean Intersection over Union (mIoU) and Panoptic Quality (PQ), each serving different segmentation goals.

\paragraph{Mean Intersection over Union (mIoU).}

The mIoU is widely used in semantic segmentation to measure the overlap between predicted and ground truth masks for each class. For a given class $c$, the $\text{IoU}_c$ is defined as

$$\text{IoU}_c = \dfrac{|O_c \cap G_c|}{|O_c \cup G_c|},$$

\noindent where $O_c$ and $G_c$ represent predicted and ground truth masks for class $c$, respectively. The mIoU score is then averaged across all classes $C$:

$$\text{mIoU} = \dfrac{1}{|C|} \sum_{c \in C} \text{IoU}_c.$$

The mIoU does not differentiate between instances of the same class. 

\paragraph{Panoptic Quality (PQ).} The PQ combines both semantic and instance segmentation and is computed as

$$
    \text{PQ}_c = 
    \underbrace{\frac{\sum_{(p,g) \in \text{TP}_c}^{} \text{IoU}(p,g)}{|\text{TP}_c|}}_\text{Segmentation Quality (SQ)}  
    \underbrace{\frac{|\text{TP}_c|}{|\text{TP}_c|\!+\!\tfrac{1}{2}|\text{FP}_c|\!+\!\tfrac{1}{2}|\text{FN}_c|}}_\text{Recognition Quality (RQ)},
$$
where $\text{TP}_c$, $\text{FP}_c$ and $\text{FN}_c$ are respectively the sets of true positives, false positives and false negatives computed after matching the predicted and ground truth instances in class $c$.
The PQ score is then averaged across all classes:

$$\text{PQ} = \frac{1}{|C|}\sum_{c \in C} \text{PQ}_c.$$

Finally, PQ\textsuperscript{$\dagger$} satisfies

$$\text{PQ}^\dagger = \dfrac{1}{|\text{things}|\!+\!|\text{stuff}|}\left(\sum_{c \in \text{things}}\text{PQ}_c +\!\!\!\sum_{c \in \text{stuff}}\text{IoU}_c\right) \!.$$

\section{Box fitting algorithm}

\begin{figure}[ht]
    \centering
    \includegraphics[width=\linewidth]{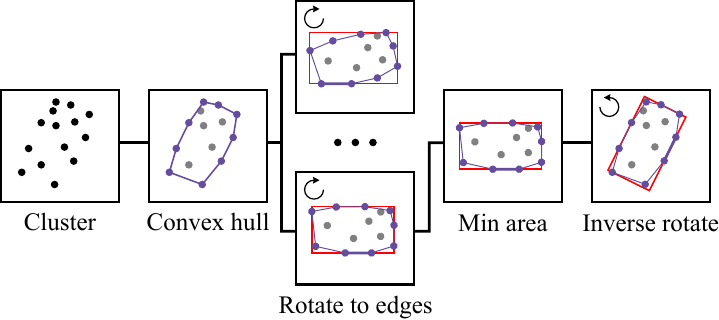}
    \caption{Visual description of the box fitting algorithm}
    \label{fig:box_fitting}
\end{figure}
\SetInd{0.25em}{0.5em} 
\begin{algorithm}
  \caption{\textbf{Box fitting algorithm.} \small This algorithm finds the best fitting box $B$ given a set of $2D$ points $P$.
  \texttt{\bfseries{convex\_hull}} returns the vertices and edges of the convex hull of \var{P}, \texttt{\bfseries{angle}} gives the angle of an edge with respect to the $x$ axis, and \texttt{\bfseries{rotate}} applies a rotation of a given angle.}
  \label{alg:box_fitting}
  \small
  \Function{fit\_box(P)}{
  \Input{Points \var{P}}
  \Output{Best fitting bounding box}
  $\var{V}, \var{E} \assign \FuncCall{convex\_hull}{P}$\;
  $\var{B}, \var{S} \assign [], []$\;
  \For{$e \in E$}{
    $\var{$\theta$} \assign \FuncCall{angle}{e}$\;
    $\var{P'} \assign \FuncCall{rotate}{P, $\boldsymbol{\theta}$}$\;
    $\var{b}^\var{x}_{\rm min} \assign \FuncCall{min}{P', axis=\var{x}}$\;
    $\var{b}^\var{x}_{\rm max} \assign \FuncCall{max}{P', axis=\var{x}}$\;
    $\var{b}^\var{y}_{\rm min} \assign \FuncCall{min}{P', axis=\var{y}}$\;
    $\var{b}^\var{y}_{\rm max} \assign \FuncCall{max}{P', axis=\var{y}}$\;
    $\var{b} \assign (\var{b}^\var{x}_{\rm min}, \var{b}^\var{y}_{\rm min}, \var{b}^\var{x}_{\rm max}, \var{b}^\var{y}_{\rm max})$\;
    $\var{S}\texttt{.append}((\var{b}[2]-\var{b}[0])*(\var{b}[3]-\var{b}[1]))$\;
    $\var{B}\texttt{.append}(\FuncCall{rotate}{b, $-\boldsymbol{\theta}$})$\;
  }
  $\var{i} \assign \FuncCall{argmin}{S}$\;
  \Return{\var{B}[i]}
  }
  \normalsize
\end{algorithm}

\vspace{-5pt}

The box fitting algorithm was borrowed from~\cite{modest} which was inspired from~\cite{stackoverflow}. It is described in \cref{alg:box_fitting} and \cref{fig:box_fitting}. It works as follows. First, we compute the convex hull of the set of points on which we need to fit a box. Second, for each edge of the convex hull, we compute its angle with respect to the x-axis, we rotate the point cloud to align this edge with the x-axis, and then fit the axis-aligned bounding box of minimum area that cover the rotated point cloud.
Finally, among all bounding boxes computed in the previous step (one for each edge of the convex hull), we keep the bounding box with the smallest area.

\section{Number of neighbors $k$.}

\begin{figure}[t]
    \vspace{-4pt}
    \centering
    \begin{tikzpicture}
        \begin{axis}[
            ybar,
            bar width=25pt,
            ymin=59,
            ymax=71,
            width=0.7\linewidth,
            height=0.4\linewidth,
            legend style={at={(0.5,-0.15)},
            anchor=north,legend columns=-1},
            ylabel={PQ},
            symbolic x coords={16, 32, 64, 128},
            xtick=data,
            nodes near coords={\pgfmathprintnumber[fixed, fixed zerofill, precision=1]{\pgfplotspointmeta}},
            nodes near coords align={vertical},
            axis x line=bottom,
            axis y line=left,
            axis line style={-},
            enlarge x limits=0.18, 
            tick label style={font=\small},
            label style={font=\small},
            legend style={font=\small},
            ]
            \addplot coordinates {(16, 65.6) (32, 65.9) (64, 66.0) (128, 66.0)};
        \end{axis}
        \node at (4.9, -0.31) {k}; 
    \end{tikzpicture}
    \vspace{-4pt}
    \caption{\textbf{Influence of the neighbors count $k$} evaluated on SemanticKITTI's validation set}
    \label{fig:ablation_k}
\end{figure}
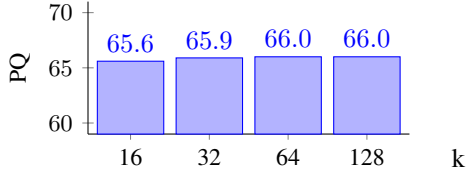

We study the variation of the performance of \ours with the choice of the number of neighbors considered in the clustering methods $k$. The results are presented in \cref{fig:ablation_k}. As mentioned in \cref{sec:clustering}, the choice of $k$ is not critical above a certain value as edges to distant neighbors will be removed by the thresholding, which we verify experimentally by finding almost no difference in PQ between $k=32$ and $64$. We thus keep $k=32$ for faster runtimes at a very little cost in terms of metrics.

\section{Choice of margins in the box splitting}

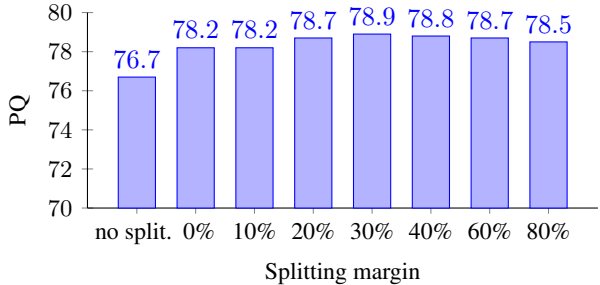
\begin{figure}[ht]
    \vspace{-4pt}
    \centering
    \begin{tikzpicture}
        \begin{axis}[
            ybar,
            bar width=14pt,
            ymin=70,
            ymax=80,
            width=\linewidth,
            height=0.5\linewidth,
            legend style={at={(0.5,-0.15)},
            anchor=north,legend columns=-1},
            ylabel={PQ},
            xlabel={Splitting margin},
            symbolic x coords={no split.\;, \;0\%, 10\%, 20\%, 30\%, 40\%, 60\%, 80\%},
            xtick=data,
            nodes near coords={\pgfmathprintnumber[fixed, fixed zerofill, precision=1]{\pgfplotspointmeta}},
            nodes near coords align={vertical},
            axis x line=bottom,
            axis y line=left,
            axis line style={-},
            enlarge x limits=0.12, 
            tick label style={font=\small},
            label style={font=\small},
            legend style={font=\small},
            ]
            \addplot coordinates {(no split.\;, 76.7) (\;0\%, 78.2) (10\%, 78.2) (20\%, 78.7) (30\%, 78.9) (40\%, 78.8) (60\%, 78.7) (80\%, 78.5)};
        \end{axis}
    \end{tikzpicture}
    \vspace{-4pt}
    \caption{\textbf{Influence of the box splitting margin} evaluated on SemanticKITTI's validation set}
    \label{fig:ablation_margin}
\end{figure}

We present in \cref{fig:ablation_margin} an ablation on the choice of the margin parameter in the box splitting algorithm. We recall that the box splitting algorithm splits clusters when they don't fit in the average box of the object class, with some tolerance (margin). The margin is expressed in $\%$ of the box size, applied in both dimensions.

First, we notice that the absence of margin hinders performance because of over-segmentation: no cluster bigger than the average box size can exist. Second, we remark that the results are stable for wide range of margins: only $0.2$ points of absolute variation in PQ for a margin between $20\%$ and $60\%$. Therefore, the margin does not need to be heavily finetuned. 

The box splitting could reach a very high number of recursions, e.g., in the pathological case where points are to be removed one by one, or a high number of iterations in the dichotomy if the border between two objects is very sensitive. For this reason, we actually put a limit in how small $dt$ can be in \cref{alg:split_cluster}. This limit is set to $1e-3$. This accelerates the algorithm by avoiding too long computations for corner cases. We verified in \cref{tab:ablation_epsilon} that this choice does not impact the performance.

\begin{table}[t]
    \centering
    \small
    \setlength{\tabcolsep}{4.1pt}
    
    \begin{tabular}{r|ccc}
    \toprule
    \multicolumn{1}{l}{$\epsilon$} & PQ & PQ\textsuperscript{$\dagger$} & RQ \\
    \midrule
    0.1     & 78.1 & 80.5 & 86.0 \\
    0.01    & 78.8 & 81.2 & 86.8 \\
    0.001   & 78.9 & 81.3 & 87.0 \\
    0.0001  & 78.9 & 81.3 & 87.0 \\
    0.00001 & 78.9 & 81.3 & 87.0 \\
    \bottomrule
    \end{tabular}
    \caption{\textbf{Influence of the box dichotomy precision limit} evaluated on nuScenes' validation set.}
    \label{tab:ablation_epsilon}
\end{table}

\section{Reporting and methodology}

In the \cref{tab:results_semkitti_val_compressed,tab:results_nuscenes_val_compressed,tab:results_semposs_scene2_compressed,tab:results_semkitti_val,tab:results_nuscenes_val,tab:results_semposs_scene2}, we decided to compare only against \ours obtained with no instance training in the semantic backbone. We reported for all methods for which we could verify, whether they used TTA or model ensembling. We checked if either the article mentioned that TTA and ensembling were not used, or the code had no mechanism in place to perform either. We put a question mark on those we could not verify.

\section{Explainability and limitations}

\ours's clustering is fully deterministic, and explainable in simple terms: in the absence of box splitting, two points $A$ and $B$ will belong to the same instance if they are predicted to be of the same semantics $c$, and there exist a path from $A$ to $B$ going only through points predicted to be of class $c$ with distance of at most $t_c$ between each edges. As such, a typical failure case exists when two instances of the same object are closer to each other than the threshold. The box splitting alleviates this issue but failures can still happen, e.g., if the distance between points within each instance is greater than the distance between the two instances. This issue vanishes as the angular resolution of the LiDAR is improved, which is experimentally verified by the higher PQ score achieved by a semantic oracle on the dense SemanticKITTI than on the sparser nuScenes.

Those weaknesses are a consequences of the simplicity and explainability of the method, and especially in the choice to use euclidean distance as the criterion of discrimination. \ours is a baseline, demonstrating what can be achieved by a training-free method, and reaches state-of-the-art results. 

\section{Per-class results}

\begin{table*}
    \small
    \centering
    \setlength{\tabcolsep}{3pt}

    \begin{tabular}{l|cc|cc|cc|cc|cc|cc|cc|cc}

    \toprule

    Method & \multicolumn{2}{c|}{\adjustbox{angle=50}{barrier}} & \multicolumn{2}{c|}{\adjustbox{angle=50}{bicycle}} & \multicolumn{2}{c|}{\adjustbox{angle=50}{bus}} & \multicolumn{2}{c|}{\adjustbox{angle=50}{car}} & \multicolumn{2}{c|}{\adjustbox{angle=50}{con. veh.}} & \multicolumn{2}{c|}{\adjustbox{angle=50}{motorcycle}} & \multicolumn{2}{c|}{\adjustbox{angle=50}{pedestrian}} & \multicolumn{2}{c}{\adjustbox{angle=50}{tra. cone}} \\
    & PQ & IoU & PQ & IoU & PQ & IoU & PQ & IoU & PQ & IoU & PQ & IoU & PQ & IoU & PQ & IoU \\

    \midrule

    \ours w. PTv3 & 61.9 & 84.4 & 77.3 & 54.0 & 83.0 & 96.1 & 93.3 & 95.2 & 64.4 & 49.8 & 89.9 & 89.7 & 93.8 & 86.6 & 92.4 & 75.0 \\
    \ours w. WI-768 & 61.0 & 84.3 & 77.5 & 54.8 & 81.0 & 94.5 & 93.2 & 95.4 & 61.1 & 49.4 & 90.8 & 90.1 & 93.6 & 86.2 & 92.6 & 76.2 \\

    Sem. Oracle & 79.5 & 100 & 95.5 & 100 & 95.9 & 100 & 97.2 & 100 & 94.9 & 100 & 97.5 & 100 & 98.1 & 100 & 99.2 & 100 \\
    
    \multicolumn{17}{l}{} \\
    \midrule
    
    Method & \multicolumn{2}{c|}{\adjustbox{angle=50}{trailer}} & \multicolumn{2}{c|}{\adjustbox{angle=50}{truck}} & \multicolumn{2}{c|}{\adjustbox{angle=50}{driveable}} & \multicolumn{2}{c|}{\adjustbox{angle=50}{other flat}} & \multicolumn{2}{c|}{\adjustbox{angle=50}{sidewalk}} & \multicolumn{2}{c|}{\adjustbox{angle=50}{terrain}} & \multicolumn{2}{c|}{\adjustbox{angle=50}{manmade}} & \multicolumn{2}{c}{\adjustbox{angle=50}{vegetation}} \\
    & PQ & IoU & PQ & IoU & PQ & IoU & PQ & IoU & PQ & IoU & PQ & IoU & PQ & IoU & PQ & IoU \\

    \midrule

    \ours w. PTv3 & 66.2 & 78.6 & 72.8 & 86.6 & 57.6 & 76.2 & 87.7 & 89.8 & 97.0 & 97.2 & 63.6 & 75.4 & 90.0 & 91.9 & 73.1 & 77.0 \\

    \ours w. WI-768 & 64.6 & 79.4 & 75.2 & 87.0 & 57.3 & 76.6 & 85.9 & 88.8 & 96.9 & 97.1 & 58.8 & 75.0 & 89.0 & 91.1 & 72.3 & 76.4 \\

    Sem. Oracle & 89.1 & 100 & 95.0 & 100 & 100 & 100 & 100 & 100 & 100 & 100 & 100 & 100 & 100 & 100 & 100 & 100 \\

    \bottomrule
    
    \end{tabular}
    \caption{Per-class panoptic segmentation results on nuScenes' validation set.}
    \label{tab:per_class_nuscenes}
\end{table*}

\begin{table*}
    \small
    \centering
    \setlength{\tabcolsep}{2.2pt}
    \begin{tabular}{l|cc|cc|cc|cc|cc|cc|cc|cc|cc|cc}

    \toprule

    Method & \multicolumn{2}{c|}{\adjustbox{angle=50}{car}} & \multicolumn{2}{c|}{\adjustbox{angle=50}{bicycle}} & \multicolumn{2}{c|}{\adjustbox{angle=50}{motorcycle}} & \multicolumn{2}{c|}{\adjustbox{angle=50}{truck}} & \multicolumn{2}{c|}{\adjustbox{angle=50}{other-vehicle}} & \multicolumn{2}{c|}{\adjustbox{angle=50}{person}} & \multicolumn{2}{c|}{\adjustbox{angle=50}{bicyclist}} & \multicolumn{2}{c|}{\adjustbox{angle=50}{motorcyclist}} & \multicolumn{2}{c|}{\adjustbox{angle=50}{road}} & \multicolumn{2}{c|}{\adjustbox{angle=50}{parking}} \\
    
    & PQ & IoU & PQ & IoU & PQ & IoU & PQ & IoU & PQ & IoU & PQ & IoU & PQ & IoU & PQ & IoU & PQ & IoU & PQ & IoU \\

    \midrule
    \ours w. WI-256 & 93.0 & 96.9 & 68.4 & 61.9 & 79.4 & 84.7 & 72.2 & 92.2 & 61.3 & 66.4 & 87.4 & 83.0 & 91.7 & 93.2 & 00.0 & 00.7 & 95.7 & 95.7 & 35.6 & 51.1 \\
    
    \ours w. MinkUNet & 93.1 & 98.0 & 69.5 & 64.8 & 81.1 & 87.0 & 79.0 & 93.7 & 73.6 & 84.9 & 87.7 & 83.3 & 93.9 & 94.3 & 12.6 & 23.4 & 94.3 & 94.7 & 42.9 & 53.7 \\
    
    Oracle & 97.4 & 100 & 94.3 & 100 & 95.0 & 100 & 97.3 & 100 & 97.5 & 100 & 98.6 & 100 & 99.5 & 100 & 99.9 & 100 & 100 & 100 & 100 & 100 \\
    
    \multicolumn{21}{l}{} \\
    \midrule
    
    Method & \multicolumn{2}{c|}{\adjustbox{angle=50}{sidewalk}} & \multicolumn{2}{c|}{\adjustbox{angle=50}{other-ground}} & \multicolumn{2}{c|}{\adjustbox{angle=50}{building}} & \multicolumn{2}{c|}{\adjustbox{angle=50}{fence}} & \multicolumn{2}{c|}{\adjustbox{angle=50}{vegetation}} & \multicolumn{2}{c|}{\adjustbox{angle=50}{trunk}} & \multicolumn{2}{c|}{\adjustbox{angle=50}{terrain}} & \multicolumn{2}{c|}{\adjustbox{angle=50}{pole}} & \multicolumn{2}{c|}{\adjustbox{angle=50}{traffic-sign}}\\
    & PQ & IoU & PQ & IoU & PQ & IoU & PQ & IoU & PQ & IoU & PQ & IoU & PQ & IoU & PQ & IoU & PQ & IoU \\

    \midrule

    \ours w. WI-256 & 81.1 & 84.4 & 00.7 & 06.3 & 89.7 & 92.6 & 29.3 & 70.0 & 87.0 & 88.3 & 61.8 & 74.7 & 59.5 & 73.7 & 63.8 & 67.3 & 60.2 & 52.6 \\

    \ours w. MinkUNet & 80.4 & 83.0 & 00.0 & 00.1 & 89.9 & 92.3 & 26.7 & 68.4 & 88.1 & 88.8 & 55.5 & 70.1 & 61.3 & 75.3 & 61.8 & 66.3 & 59.0 & 49.9 \\

    Oracle & 100 & 100 & 100 & 100 & 100 & 100 & 100 & 100 & 100 & 100 & 100 & 100 & 100 & 100 & 100 & 100 & 100 & 100 \\

    \bottomrule
    
    \end{tabular}
    \caption{Per-class panoptic segmentation results on SemanticKITTI' validation set.}
    \label{tab:per_class_semantickitti}
\end{table*}

For completeness, we present in \cref{tab:per_class_nuscenes,tab:per_class_semantickitti} the per-class results of our method on both nuScenes and SemanticKITTI. 

\section{Performance of \ours and distance to the sensor}

\begin{table}[ht]
    \small
    \centering
    \setlength{\tabcolsep}{2pt}
    \captionsetup{font=footnotesize}
\resizebox{\columnwidth}{!}{%
    \begin{tabular}{l|ccc|ccc|ccc}
    
    & \multicolumn{3}{c|}{0-15m} & \multicolumn{3}{c|}{15-30m} & \multicolumn{3}{c}{30m+} \\
    
     Method & PQ & PQ\textsuperscript{$\dagger$} & mIoU & PQ & PQ\textsuperscript{$\dagger$} & mIoU & PQ & PQ\textsuperscript{$\dagger$} & mIoU
    \\
    
    \midrule
    \ours w/ Ens.       & 68.5 & 71.0 & 73.0 & 68.6 & 70.7 & 68.4 & 65.8 & 65.9 & 54.4 \\
    P3Former            & 63.6 & 65.8 & 67.8 & 64.8 & 66.9 & 63.4 & 59.6 & 60.1 & 47.2 \\
    P3Former \& \ours   & 64.0 & 66.2 & 67.8 & 64.8 & 66.9 & 63.4 & 59.4 & 59.8 & 47.2 \\
    
    \end{tabular}}
    \caption{\textbf{Performance w.r.t sensor distance} on SemanticKITTI's validation set.}
    \label{tab:distance}
\end{table}

We report in \cref{tab:distance} a breakdown of \ours's performance with respect to the distance to the sensor. \ours improves over P3Former at short range, but slightly deteriorates at long range. This suggests that \ours is particularly efficient on dense point clouds (i.e. close to the sensor) but gets lesser performance on sparser point clouds.

\section{More visualisations}

\begin{figure*}[t]
    \centering

    \setlength{\tabcolsep}{3pt}
    \begin{tabular}{@{}c|c|c@{}}
    \includegraphics[trim={0cm 3cm 0cm 0cm},clip,width=0.32\linewidth]{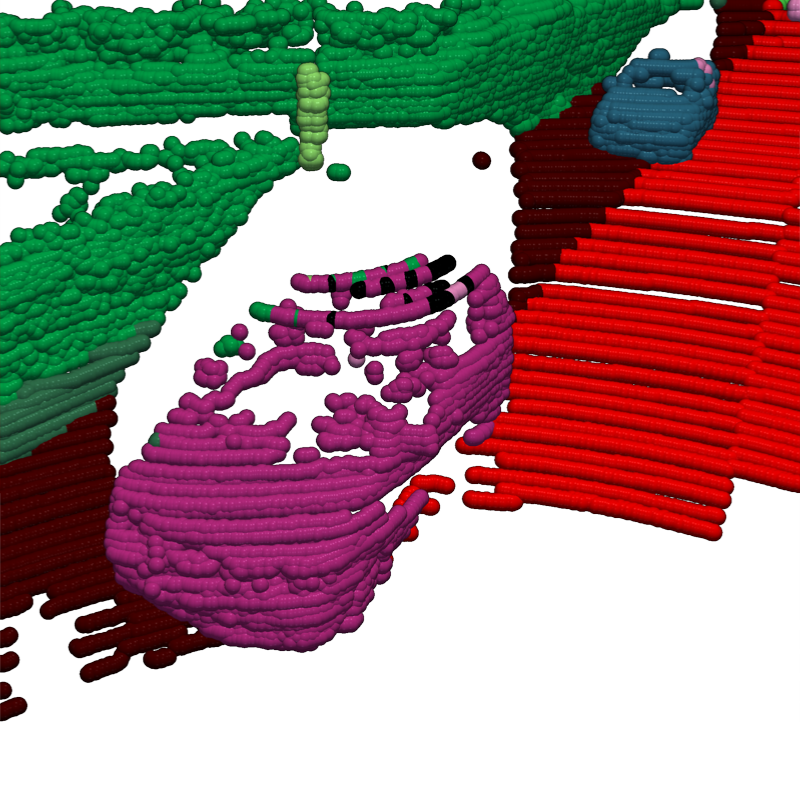}
    &
    \includegraphics[trim={0cm 3cm 0cm 0cm},clip,width=0.32\linewidth]{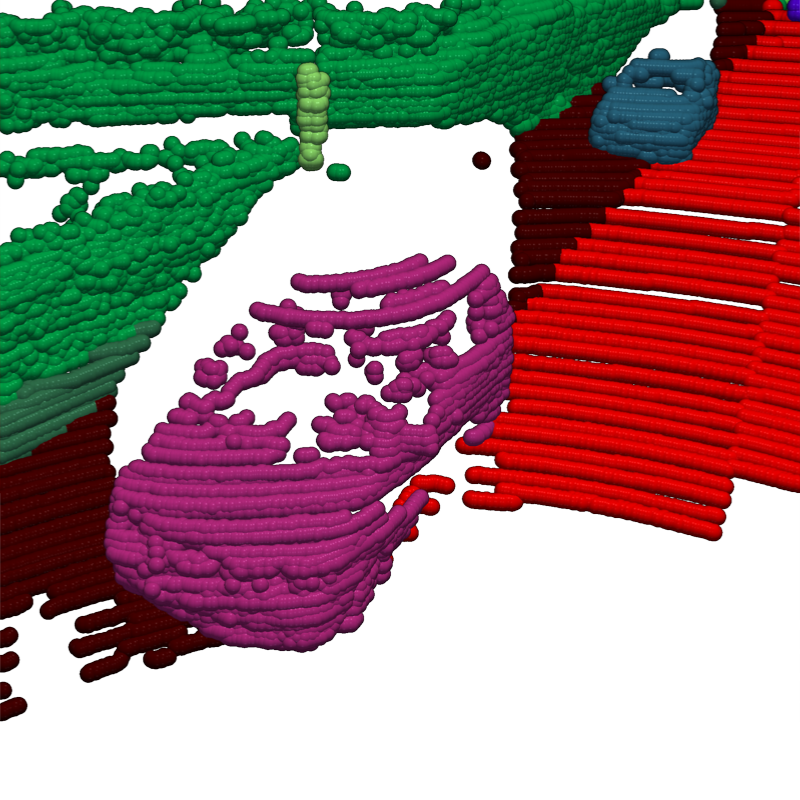}
    &
    \includegraphics[trim={0cm 3cm 0cm 0cm},clip,width=0.32\linewidth]{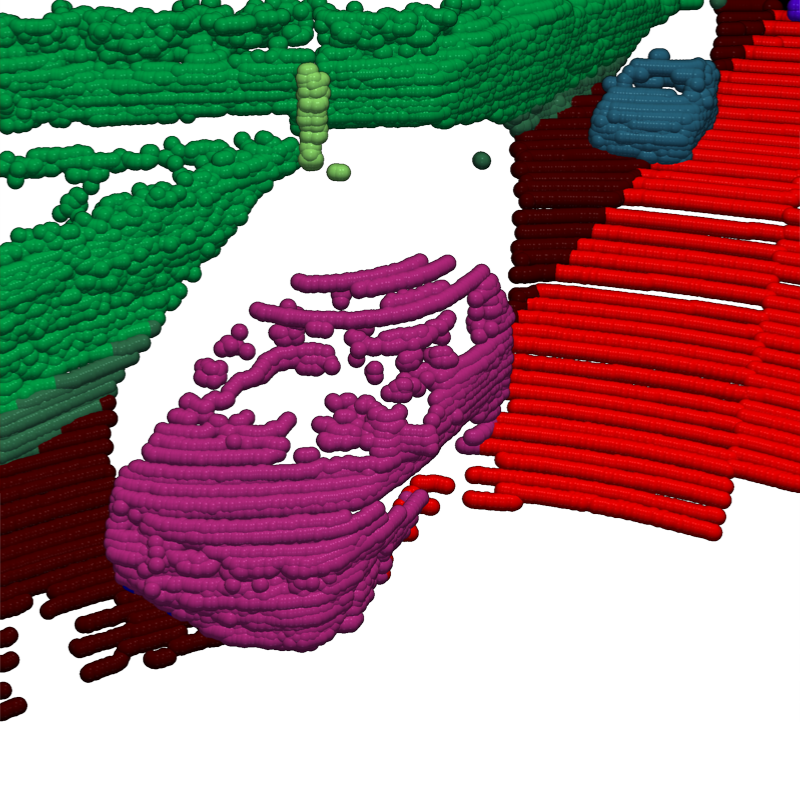}
    \\
    \includegraphics[trim={0cm 0cm 0cm 5cm},clip,width=0.32\linewidth]{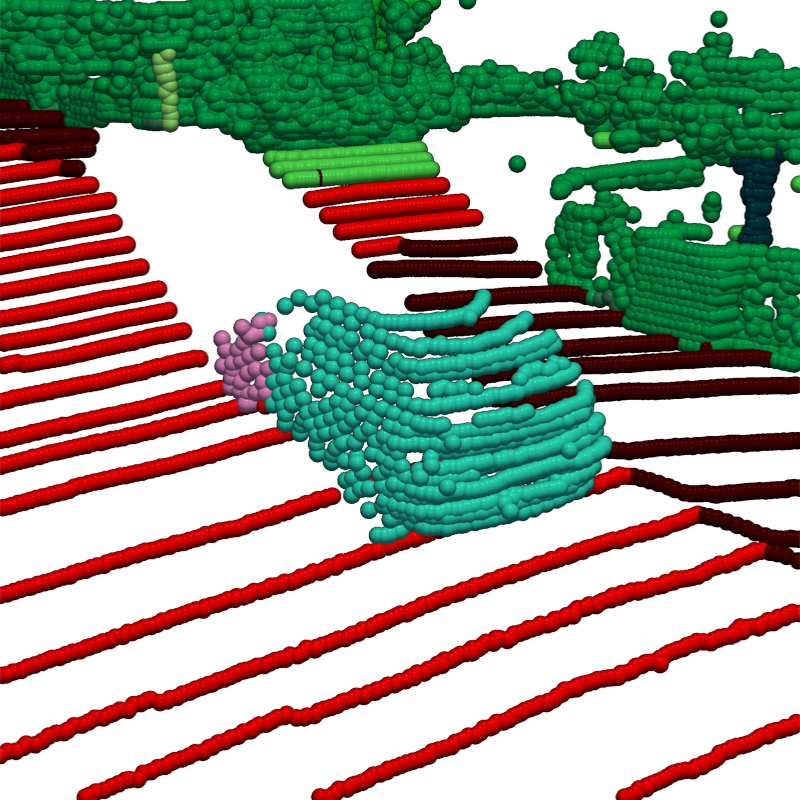}
    &
    \includegraphics[trim={0cm 0cm 0cm 5cm},clip,width=0.32\linewidth]{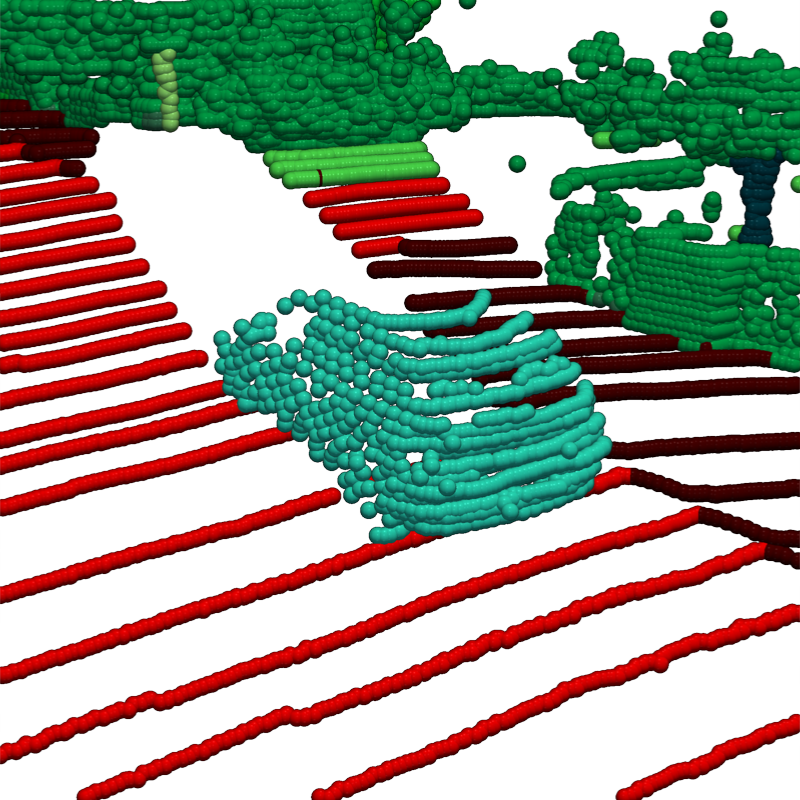}
    &
    \includegraphics[trim={0cm 0cm 0cm 5cm},clip,width=0.32\linewidth]{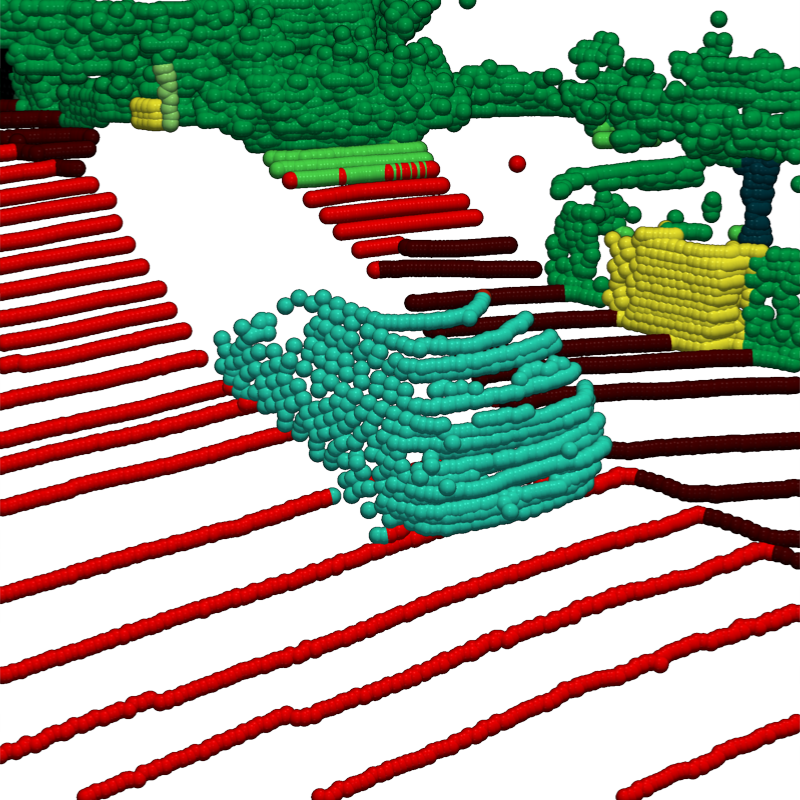}
    \\
    \includegraphics[trim={0cm 0cm 0cm 2cm},clip,width=0.32\linewidth]{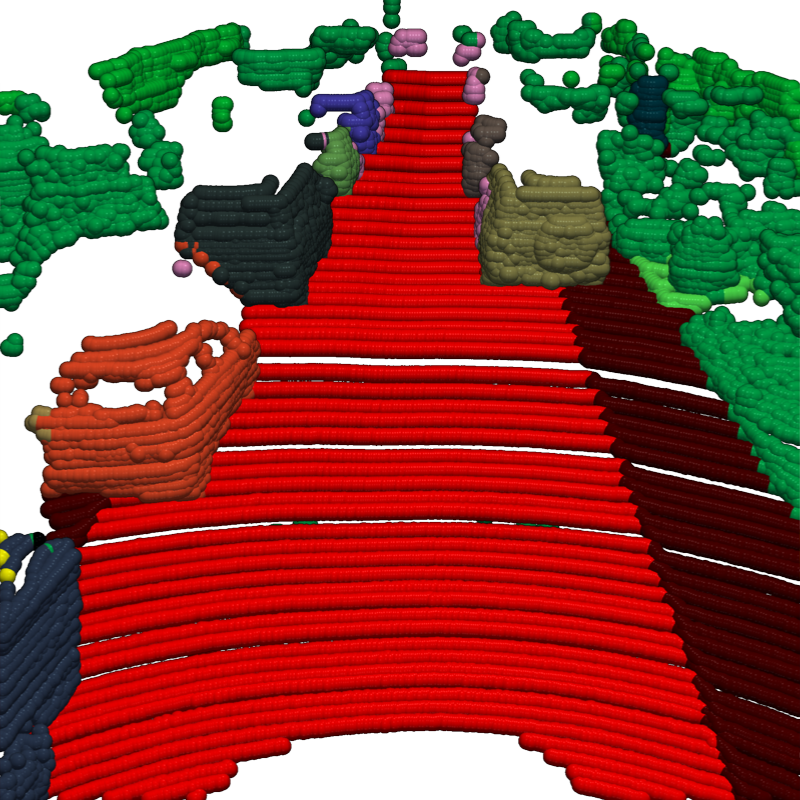}
    &
    \includegraphics[trim={0cm 0cm 0cm 2cm},clip,width=0.32\linewidth]{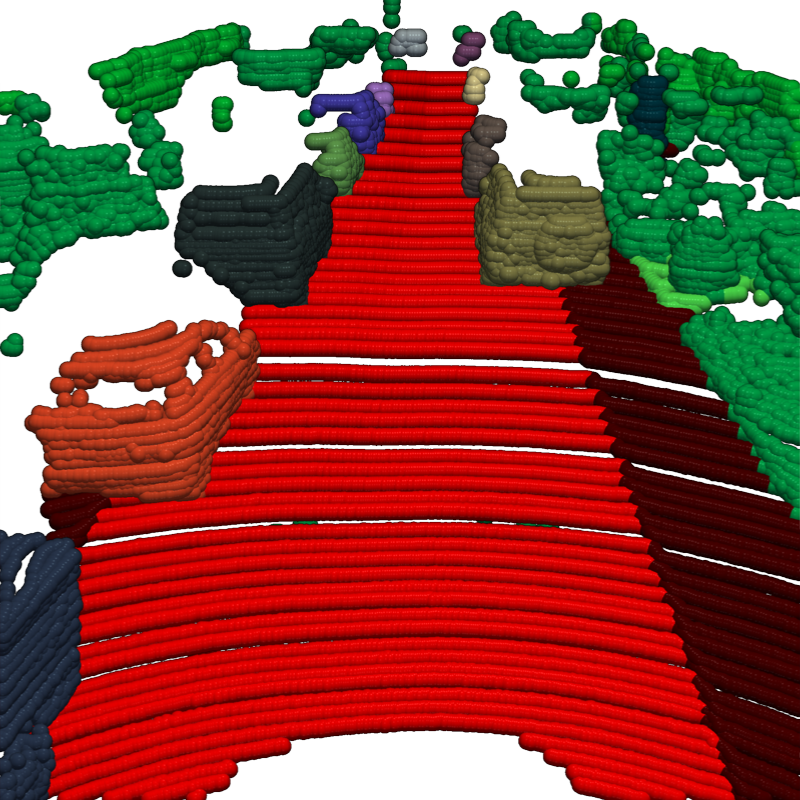}
    &
    \includegraphics[trim={0cm 0cm 0cm 2cm},clip,width=0.32\linewidth]{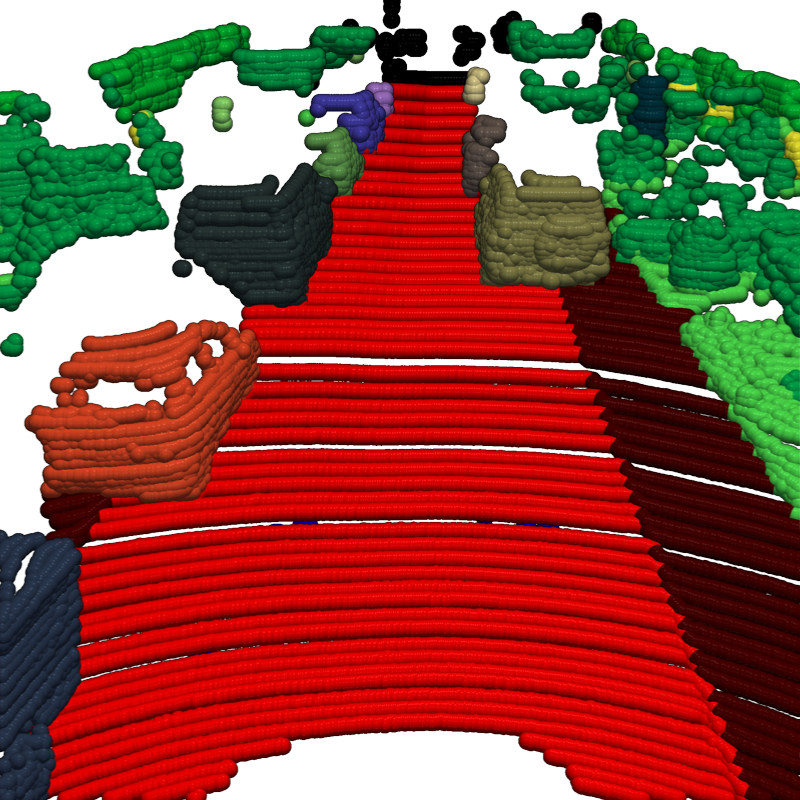}
    \\
    \includegraphics[trim={0cm 2cm 0cm 0cm},clip,width=0.32\linewidth]{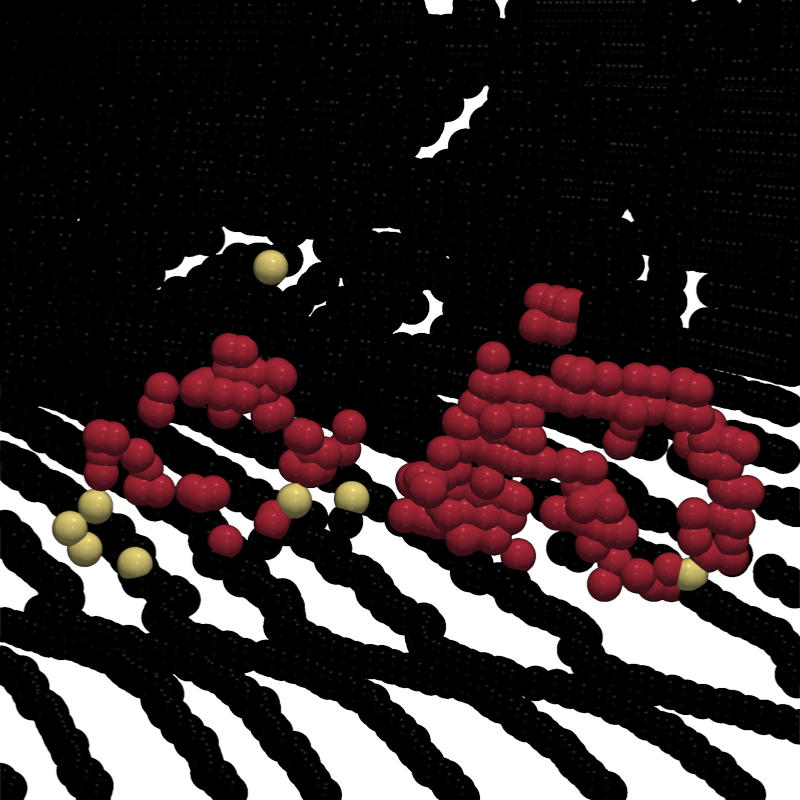}
    &
    \includegraphics[trim={0cm 2cm 0cm 0cm},clip,width=0.32\linewidth]{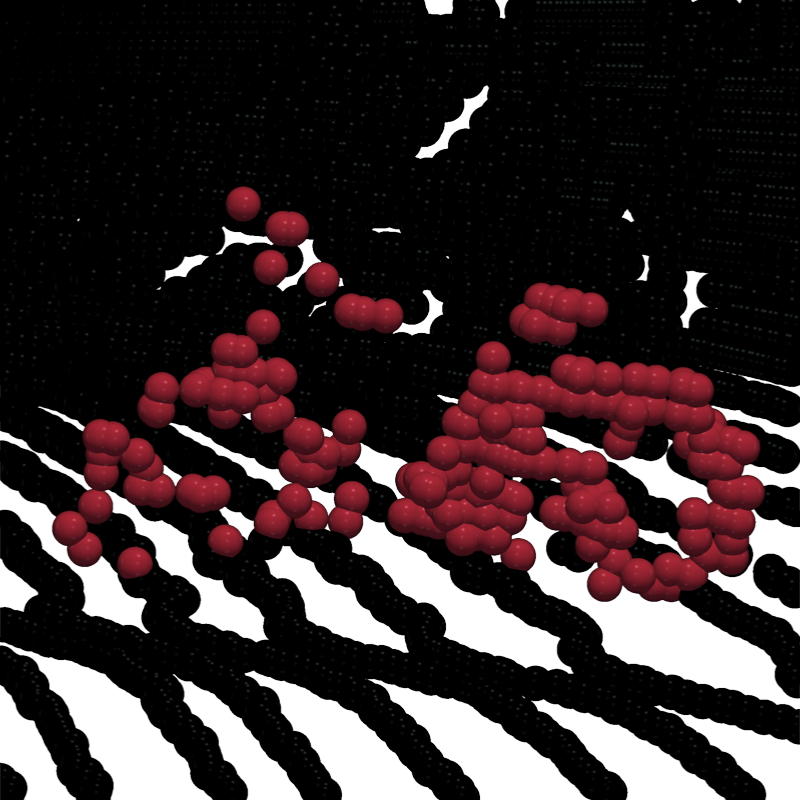}
    &
    \includegraphics[trim={0cm 2cm 0cm 0cm},clip,width=0.32\linewidth]{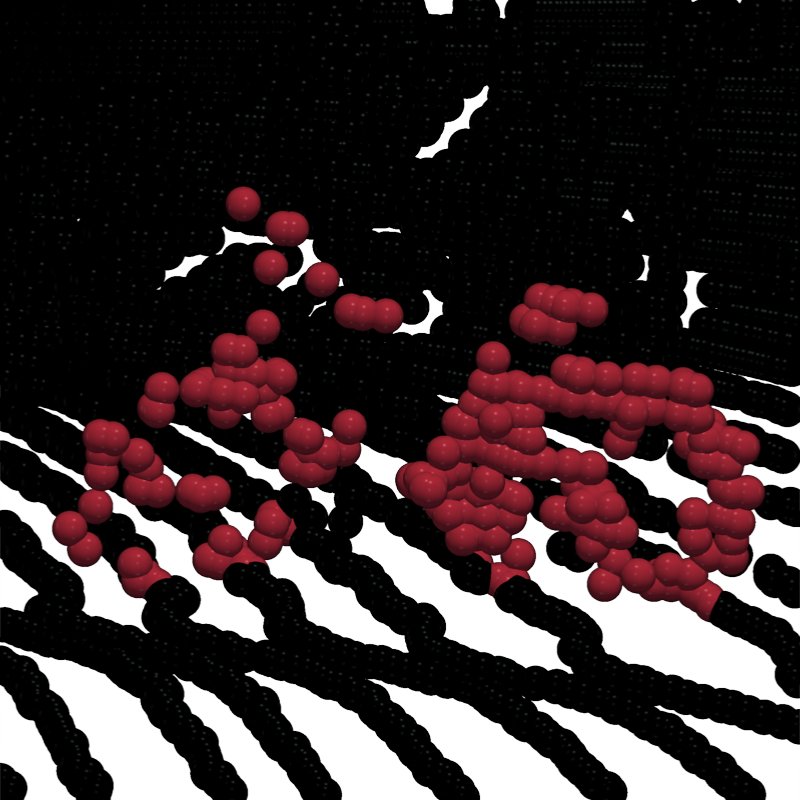}
    \\
    D\&M~\cite{dividemerge} w.~MinkUNet & \ours w.~MinkUNet & Ground Truth
    \end{tabular}

    \caption{\textbf{Examples of panoptic predictions on SemanticKITTI.} We present the results obtained with D\&M~\cite{dividemerge} (left) and \ours (middle). Both methods use the same MinkUNet to obtain pointwise semantic predictions. The Ground Truth masks are presented on the rightmost panels. On the last row, only points predicted as ``bicycle'' are colored for visual clarity.}
    \label{fig:more_visualizations}
\end{figure*}

More visualization of our method can be seen on \cref{fig:more_visualizations}. We can see that D\&M has some boundaries issues, which \ours does not have. Furthermore, on those examples, the remaining false predictions are mostly semantic prediction errors from the MinkUNet, rather than errors in our clustering.

\begin{table*}[ht]
    \small 
    \centering
    \setlength{\tabcolsep}{3pt}
    \begin{tabular}{ll|ccc|accc|ccc|ccc|c}
    
    \toprule
    
    Method &&Inst.& TTA & Ens. & PQ & PQ\textsuperscript{$\dagger$} & RQ & SQ & PQ\textsubscript{Th} & RQ\textsubscript{Th} & SQ\textsubscript{Th} & PQ\textsubscript{St} & RQ\textsubscript{St} & SQ\textsubscript{St} & mIoU \\
    && labels & \\

    \midrule

    
    
    LPSAD (impl. from \cite{gps3net}) & \cite{lidarpanoptic}    & \cmark & \qmark & \qmark &
    37.4 & 44.2 & 47.8 & 66.9 & 25.3 & 32.4 & 65.2 & 46.2 & 58.9 & 68.2 & 49.4 \\ 

    PanopticTrackNet           & \cite{mopt}                    & \cmark & \qmark & \qmark &
    40.0 & -    & 48.3 & 73.0 & 29.9 & 33.6 & 76.8 & 47.4 & 59.1 & 70.3 & 53.8 \\ 

    
    PointGroup                 & \cite{pointgroup}             & \cmark & \xmark & \xmark &
    46.1 & 54.0 & 56.6 & 74.6 & 47.7 & 55.9 & 73.8 & 45.0 & 57.1 & 75.1 & 55.7 \\ 




    
    TORNADO-Net (fusion)       & \cite{tornadonet}         & \cmark & \qmark & \qmark &
    50.6 & 55.9 & 62.1 & 74.9 & 48.1 & 57.5 & 72.5 & 52.4 & 65.4 & 76.7 & 59.2 \\ 


    
    Panoster                   & \cite{panoster}            & \cmark & \xmark & \xmark &
    55.6 & -    & 66.8 & 79.9 & 56.6 & 65.8 & -    & -    & -    & -    & 61.1 \\ 

    LCPS (lidar)               & \cite{lcps}               & \cmark & \xmark & \xmark &
    55.7 & 65.2 & 65.8 & 74.0 & -    & -    & -    & -    & -    & -    & 61.1 \\ 
    
    Cylinder3D \& D\&M         & \cite{dividemerge}         & \xmark & \xmark & \xmark &
    57.2 & -    & -    & -    &  -    & -    & -    & -    & -    & -    & - \\ 
    

    Location-Guided            & \cite{locationguided}     &  \cmark & \qmark & \qmark &
    59.0 & 63.1 & 69.4 & 78.7 & 65.3 & 73.5 & 88.5 & 53.9 & 66.4 & 71.6 & 61.4 \\ 
    
    Panoptic-PolarNet          & \cite{panopticpolarnet}    & \cmark & \xmark & \xmark &
    59.1 & 64.1 & 70.2 & 78.3 & 65.7 & 74.7 & 87.4 & 54.3 & 66.9 & 71.6 & 64.5 \\ 
    
    EfficientLPS               & \cite{efficientlps}        & \cmark & \xmark & \xmark &
    59.2 & 65.1 & 69.8 & 75.0 & 58.0 & 68.2 & 78.0 & 60.9 & 71.0 & 72.8 & 64.9 \\ 
    
    MaskPLS                    & \cite{maskpls}            & \cmark & \xmark & \xmark &
    59.8 & -    & 69.0 & 76.3 & -    & -    & -    & -    & -    & -    & 61.9 \\ 

    DS-Net (SPVCNN)            & \cite{dsnet2}             & \cmark & \qmark & \qmark &
    61.4 & 65.2 & 72.7 & 79.0 & 65.2 & 72.3 & 79.3 & 57.9 & 71.1 & 79.3 & 69.6 \\ 
    
    Panoptic-PHNet             & \cite{phnet}               & \cmark & \xmark & \xmark &
    61.7 & -    & -    & -    & 69.3 & -    & -    & -    & -    & -    & 65.7 \\ 
    
    PANet                      & \cite{panet}               & \cmark & \qmark & \qmark &
    61.7 & 66.6 & 71.8 & 79.3 & -    & -    & -    & -    & -    & -    & 68.1 \\ 

    D\&M w. MinkUnet & \cite{dividemerge} & \xmark & \mmark & \xmark & 
    61.8 & 66.2 & 72.8 & 79.6 & 64.5 & 73.7 & 87.2 & 59.9 & 72.1 & 74.2 & 71.4\\
    
    CenterLPS                  & \cite{centerlps}         & \cmark & \qmark & \qmark & 
    62.1 & 67.0 & 72.0 & 80.7 & 68.4 & 75.2 & 91.0 & 57.5 & 69.7 & 73.2 & 68.1 \\ 

    \rowcolor{green!15}
    \ours w.~PTv3            &                          & \xmark & \mmark & \xmark & 
    62.4 & 66.2 & 72.0 & 76.7 & 66.1 & 72.1 & 80.0 & 59.8 & 71.9 & 74.3 & 67.3 \\

    P3Former                   & \cite{p3former}          & \cmark & \xmark & \xmark & 
    62.6 & 66.2 & 72.4 & 76.2 & 69.4 & -    & -    & 57.7 & -    & -    & -    \\
    
    CFNet                      & \cite{cfnet}        & \cmark & \xmark & \xmark & 
    62.7 & 67.5 & -    & -    & 70.0 & -    & -    & 57.3 & -    & -    & 67.4 \\ 
    
    LPST                       & \cite{lpst}                & \cmark & \xmark & \xmark & 
    63.1 & 70.8 & 73.1 & 79.2 & 68.7 & 75.7 & 86.7 & 58.9 & 71.2 & 73.7 & 69.7 \\ 
    
    GP-S3Net              & \cite{gps3net}           & \cmark & \qmark & \qmark & 
    63.3 & 71.5 & 75.9 & 81.4 & 70.2 & 80.1 & 86.2 & 58.3 & 72.9 & 77.9 & 73.0 \\ 
    
    DQFormer                   & \cite{dqformer}           & \cmark & \xmark & \xmark &
    63.5 & 67.2 & 73.1 & 81.7 & -    & -    & -    & -    & -    & -    & - \\ 
    
    \rowcolor{green!15}
    \ours w.~MinkUNet    &                          & \xmark & \xmark & \xmark & 
    64.2 & 68.9 & 74.1 & 84.4 & 72.6 & 79.3 & 90.6 & 58.1 & 70.4 & 79.8 & 70.7 \\
    
    \rowcolor{green!15}
    \ours w.~WI-256            &                          & \phantom{*}\xmark* & \mmark & \xmark & 
    64.2 & 69.0 & 74.1 & 79.7 & 69.3 & 75.9 & 79.8 & 60.4 & 72.7 & 79.7 & 70.3 \\

    PUPS (w/o ens.)            & \cite{pups}              & \cmark & \xmark & \xmark & 
    64.4 & 68.6 & 74.1 & 81.5 & 73.0 & 79.3 & 92.6 & 58.1 & 70.4 & 73.5 & - \\ 

    IEQLPS                     & \cite{ieqlps}            & \cmark & \qmark & \qmark & 
    64.7 & 68.1 & 74.7 & 81.3 & 73.5 & 79.1 & 93.0 & 58.3 & 71.4 & 72.8 & - \\ 

    \rowcolor{green!15}
    \ours w.~MinkUNet         &                          & \xmark & \mmark & \xmark & 
    65.9 & 70.2 & 75.5 & 81.4 & 73.9 & 80.1 & 91.2 & 60.0 & 72.2 & 74.2 & 72.2 \\
    
    PUPS              & \citep{pups}             & \cmark & \cmark & \cmark & 
    66.3 & 70.2 & 75.6 & 82.5 & 74.6 & 80.3 & 93.4 & 60.2 & 72.2 & 74.5 & - \\ 
    
    \rowcolor{green!15}
    \ours w. MinkUNet \& PTv3 & 
    & \xmark & \mmark & \mmark &  
    66.6 & 70.8 & 76.1 & 82.6 & 73.4 & 79.5 & 93.2 & 61.6 & 73.6 & 74.9 & 72.0 \\

    \bottomrule

    \multicolumn{16}{l}{*: the publicly available model for WI-256~\cite{waffleiron} used instance annotations in its data augmentation pipeline}\\
    \end{tabular}
    \caption{Panoptic segmentation results on the validation set of SemanticKITTI. `TTA' and `Ens.' stand for Test-Time Augmentation and Ensembling. \mmark{} denotes that TTA/ensemble was used on the semantic head only. The main metric is the PQ. }
    \label{tab:results_semkitti_val}
\end{table*}

\begin{table*}
    \small
    \centering
    \setlength{\tabcolsep}{3pt}

    \begin{tabular}{ll|ccc|accc|ccc|ccc|c}

    \toprule
    
    Method &&Inst.& TTA & Ens. & PQ & PQ\textsuperscript{$\dagger$} & RQ & SQ & PQ\textsubscript{Th} & RQ\textsubscript{Th} & SQ\textsubscript{Th} & PQ\textsubscript{St} & RQ\textsubscript{St} & SQ\textsubscript{St} & mIoU \\
    && labels & \\

    \midrule
    
    
    
    
    LPSAD (impl. from \cite{gps3net}) & \cite{lidarpanoptic}    & \cmark & \qmark & \qmark & 
    50.4 & 57.7 & 62.4 & 79.4 & 43.2 & 53.2 & 80.2 & 57.5 & 71.7 & 78.5 & 62.5 \\ 
    
    
    PanopticTrackNet  & \cite{mopt}                             & \cmark & \qmark & \qmark & 
    51.4 & 56.2 & 63.3 & 80.2 & 45.8 & 55.9 & 81.4 & 60.4 & 75.5 & 78.3 & 58.0 \\ 
    
    VIN               & \cite{vin}                              & \cmark & \xmark & \qmark & 
    51.7 & 57.4 & 61.8 & 82.6 & 45.7 & 53.7 & 83.6 & 61.8 & 75.4 & 80.9 & 73.7 \\ 
    
    TORNADO-Net (fusion)  & \cite{tornadonet}                   & \cmark & \qmark & \qmark & 
    54.0 & 59.8 & 65.4 & 80.9 & 44.1 & 53.9 & 80.1 & 63.9 & 76.9 & 81.8 & 68.0 \\ 
    
    MaskPLS           & \cite{maskpls}                          & \cmark & \xmark & \xmark & 
    57.7 & 60.2 & 66.0 & 71.8 & 64.4 & 73.3 & 84.8 & 52.2 & 60.7 & 62.4 & 62.5 \\ 
    
    
    
    
    GP-S3Net          & \cite{gps3net}                          & \cmark & \qmark & \qmark & 
    61.0 & 67.5 & 72.0 & 84.1 & 56.0 & 65.2 & 85.3 & 66.0 & 78.7 & 82.9 & 75.8 \\ 
    
    EfficientLPS      & \cite{efficientlps}                     & \cmark & \xmark & \xmark & 
    62.0 & 65.6 & 73.9 & 83.4 & 56.8 & 68.0 & 83.2 & 70.6 & 83.6 & 83.8 & 65.6 \\ 
    
    DS-Net (SPVCNN)   & \cite{dsnet2}                            & \cmark & \qmark & \qmark & 
    64.7 & 67.6 & 76.1 & 83.5 & 58.6 & 64.2 & 82.8 & 74.7 & 86.5 & 85.5 & 76.3 \\ 
    
    PVCL              & \cite{pvcl}                             & \cmark & \qmark & \qmark & 
    64.9 & 67.8 & 77.9 & 81.6 & 59.2 & 72.5 & 79.7 & 67.6 & 79.1 & 77.3 & 73.9 \\ 
    
    SCAN              & \cite{scan}                             & \cmark & \qmark & \qmark & 
    65.1 & 68.9 & 75.3 & 85.7 & 60.6 & 70.2 & 85.7 & 72.5 & 83.8 & 85.7 & 77.4 \\ 
    
    Panoptic-PolarNet & \cite{panopticpolarnet}                 & \cmark & \xmark & \xmark & 
    67.7 & 71.0 & 78.1 & 86.0 & 65.2 & 74.0 & 87.2 & 71.9 & 84.9 & 83.9 & 69.3 \\ 
    
    SMAC-Seg          & \cite{smacseg}                          & \cmark & \qmark & \qmark & 
    68.4 & 73.4 & 79.7 & 85.2 & 68.0 & 77.2 & 87.3 & 68.8 & 82.1 & 83.0 & 71.2 \\ 
    
    PANet             & \cite{panet}                            & \cmark & \xmark & \xmark & 
    69.2 & 72.9 & 80.7 & 85.0 & 69.5 & 79.3 & 86.7 & 68.7 & 82.9 & 82.1 & 72.6 \\ 
    
    CPSeg             & \cite{cpseg}                            & \cmark & \qmark & \qmark & 
    71.1 & 75.6 & 82.5 & 85.5 & 71.5 & 81.3 & 87.3 & 70.6 & 83.7 & 83.6 & 73.2 \\ 
    
    LCPS (lidar)      & \cite{lcps}                             & \cmark & \xmark & \xmark & 
    72.9 & 77.6 & 82.0 & 88.4 & 72.8 & 80.5 & 90.1 & 73.0 & 84.5 & 85.5 & 75.1 \\ 
    
    PUPS              & \cite{pups}                             & \cmark & \qmark & \qmark & 
    74.7 & 77.3 & 83.3 & 89.4 & 75.4 & 81.9  &91.8 & 73.6 & 85.6 & 85.3 & - \\
    
    Panoptic-PHNet    & \cite{phnet}                            & \cmark & \xmark & \xmark & 
    74.7 & 77.7 & 84.2 & 88.2 & 74.0 & 82.5 & 89.0 & 75.9 & 86.9 & 86.8 & 79.7 \\ 
    
    CFNet             & \cite{cfnet}                            & \cmark & \xmark & \xmark & 75.1 & 78.0 & 84.6 & 88.8 & 74.8 & 82.9 & 89.8 & 76.6 & 87.3 & 87.1 & 79.3 \\
    
    P3Former          & \cite{p3former}                         & \cmark & \xmark & \xmark & 
    75.9 & 78.9 & 84.7 & 89.7 & 76.9 & 83.3 & 92.0 & 75.4 & 87.1 & 86.0 & - \\ 
    
    CenterLPS         & \cite{centerlps}                        & \cmark & \qmark & \qmark & 
    76.4 & 79.2 & 86.2 & 88.0 & 77.5 & 87.1 & 88.4 & 74.6 & 84.9 & 87.3 & 77.1\\ 

    \rowcolor{green!15}
    \ours w.~WI-768  & & \xmark & \xmark & \xmark & 
    76.9 & 79.9 & 85.7 & 89.3 & 77.9 & 85.3 & 90.9 & 75.3 & 86.3 & 86.7 & 80.3 \\
    
    IEQLPS            & \cite{ieqlps}                           & \cmark & \qmark & \qmark & 
    77.1 & 79.1 & 86.5 & 88.2 & 79.5 & 86.6 & 91.7 & 73.0 & 86.4 & 83.9 & - \\ 

    LPST               & \cite{lpst}                            & \cmark & \xmark & \xmark & 
    77.1 & 79.9 & 86.5 & 88.6 & 79.3 & 87.5 & 90.3 & 73.6 & 84.9 & 85.7 & 80.3 \\ 
    
    
    DQFormer          & \cite{dqformer}                         & \cmark & \xmark & \xmark & 
    77.7 & 79.5 & 86.8 & 89.2 & 77.8 & 86.7 & 89.5 & 77.5 & 87.0 & 88.6 & - \\ 

    
    \rowcolor{green!15}
    \ours  w.~WI-768  & & \xmark & \mmark & \xmark &
    77.9 & 80.7 & 86.5 & 89.6 & 78.6 & 86.0 & 91.0 & 76.7 & 87.3 & 87.3 & 81.4\\
    
    \rowcolor{green!15}
    \ours w.~PTv3    & & \xmark & \mmark & \xmark & 
    78.9 & 81.3 & 87.0 & 90.4 & 79.3 & 86.2 & 91.7 & 78.2 & 88.3 & 88.1 & 81.5\\
    
    \rowcolor{green!15}
    \ours w. WI-768 \& PTV3 & & \xmark & \mmark & \mmark & 
    79.5 & 81.9 & 87.6 & 90.5 & 80.1 & 87.1 & 91.7 & 78.6 & 88.5 & 88.3 & 82.7 \\

    LidarMultiNet     & \cite{lidarmultinet}                    & \cmark & \cmark & \cmark & 81.8 & -    & 89.7 & 90.8 & -    & -    & -    & -    & -    & -    & 83.6 \\ 

    \bottomrule
    
    \end{tabular}
    \caption{Panoptic segmentation results on the validation set of nuScenes. `TTA' and `Ens.' stand for Test-Time Augmentation and Ensembling. \mmark{} denotes that TTA/ensemble was used on the semantic head only. The main metric is the PQ.}
    \label{tab:results_nuscenes_val}
\end{table*}

\begin{table*}[t]

    \small
    \centering
    \setlength{\tabcolsep}{3pt}
    
    \begin{tabular}{ll|ccc|accc|ccc|ccc|c}

    \toprule
    
    Method &&Inst.& TTA & Ens. & PQ & PQ\textsuperscript{$\dagger$} & RQ & SQ & PQ\textsubscript{Th} & RQ\textsubscript{Th} & SQ\textsubscript{Th} & PQ\textsubscript{St} & RQ\textsubscript{St} & SQ\textsubscript{St} & mIoU \\
    && labels & \\

    \midrule

    LPSAD (impl. from \cite{gps3net}) & \cite{lidarpanoptic}& \cmark & \qmark & \qmark & 
    22.5 & 32.7 & 34.0 & 53.5 & 18.7 & 25.7 & 70.5 & 24.0 & 37.1 & 47.1 & 35.5\\ 
    TORNADO-Net (fusion)                  & \cite{tornadonet}   & \cmark & \qmark & \qmark & 
    33.7 & 43.3 & 46.0 & 68.4 & 41.2 & 49.6 & 83.1 & 30.9 & 44.7 & 62.9 & 44.5 \\ 
    DS-Net                                & \cite{dsnet}        & \cmark & \qmark & \qmark & 
    35.6 & 45.9 & 49.2 & 68.6 & 27.4 & 33.8 & 76.8 & 38.7 & 55.0 & 65.5 & 54.5 \\ 
    GP-S3Net                              & \cite{gps3net}      & \cmark & \qmark & \qmark & 
    48.7 & 60.3 & 63.7 & 61.3 & 61.6 & 71.7 & 86.4 & 43.8 & 60.8 & 51.8 & 61.8 \\ 
    \rowcolor{green!15}
    \ours w.~PTv3                        &              & \xmark & \xmark & \xmark & 
    51.4 & 57.7 & 67.2 & 74.3 & 64.8 & 74.1 & 87.2 & 47.4 & 65.1 & 70.5 & 58.3 \\ 

    \bottomrule
    
    \end{tabular}
    \caption{Panoptic segmentation results on Sequence $02$ of SemanticPOSS as validation set.}
    \label{tab:results_semposs_scene2}
\end{table*}

\end{document}